%% file: neurips_2025.tex
\documentclass{article}

 \usepackage[preprint]{neurips_2025}

\usepackage[utf8]{inputenc} 
\usepackage[T1]{fontenc}    
\usepackage{hyperref}       
\usepackage{url}            
\usepackage{booktabs}       
\usepackage{amsfonts}       
\usepackage{nicefrac}       
\usepackage{microtype}      
\usepackage{xcolor}         
\usepackage{graphicx}
\usepackage{amsmath}
\usepackage{wrapfig}
\usepackage{dsfont}
\usepackage{tabularx}
\usepackage[table]{xcolor}

\newcommand{\MyTitleWidth}{1.12\textwidth} 

\title{\texorpdfstring{%
  \makebox[\textwidth][c]{%
    \parbox{\MyTitleWidth}{\centering
      Thinking with Blueprints: Assisting Vision-Language~Models\\
      in Spatial Reasoning via Structured Object Representation}%
  }%
}{Thinking with Blueprints: Assisting Vision--Language Models in Spatial Reasoning via Structured Object Representation}}

\newcommand{\corrauthor}{\thanks{Corresponding author.}}
\newcommand{\corrauthormark}{\addtocounter{footnote}{-1}\footnotemark}
\author{%
  Weijian Ma\thanks{Work done during internship at Microsoft Research, Asia.} \\
  National University of Singapore \\
  \texttt{weijian.ma@u.nus.edu} \\
  \And
  Shizhao Sun\corrauthor  \\
  Microsoft Research, Asia \\
  \texttt{shizsu@microsoft.com} \\
  \And
  Tianyu Yu \\
  Tsinghua University \\
  \texttt{yiranytianyu@gmail.com}
  \And
  Ruiyu Wang \\
  University of Toronto \\
  \texttt{rwang@cs.toronto.edu} \\
  \And
  Tat-Seng Chua\corrauthormark \\
  National University of Singapore \\
  \texttt{dcscts@nus.edu.sg} \\
  \And  
  Jiang Bian \\
  Microsoft Research, Asia \\
  \texttt{jiabia@microsoft.com} \\
}

\begin{document}

\maketitle

\input{sec/0_abstract} 
\begin{figure}
    \centering
    \includegraphics[width=\linewidth]{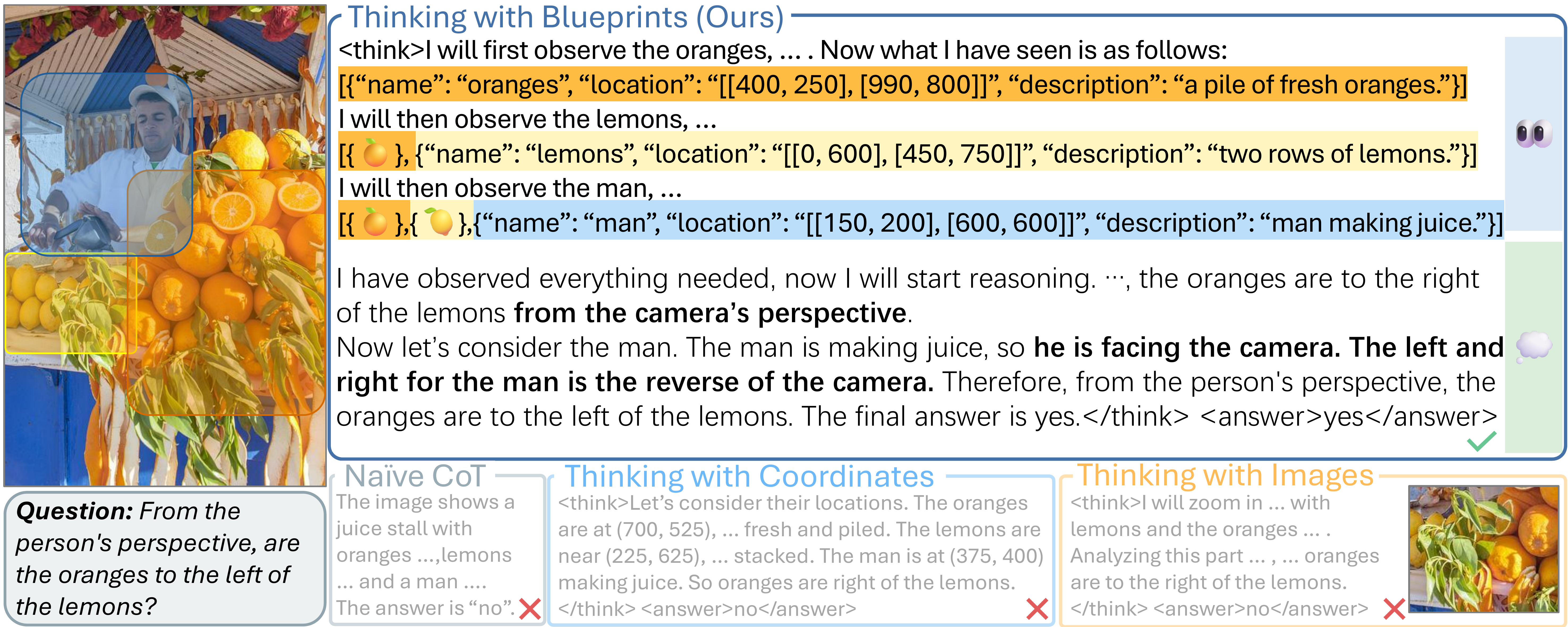}
    \caption{
        An illustrative comparison of our method with other spatial reasoning approaches.
        Inspired by the cognitive concept of an object-centric blueprint, our method first constructs a JSON-style blueprint recording the positions, sizes, and attributes of relevant objects, and then reasons over this structured representation to produce the final answer.
        Other approaches overlook such an explicit and global blueprint during reasoning, often resulting in superficial analysis and incorrect answers.
    }
    \label{fig:overall_illustration}
\end{figure}
\input{sec/1_intro}
\input{sec/2_related_work}
\begin{figure*}[t]
    \centering
    \includegraphics[width=\linewidth]{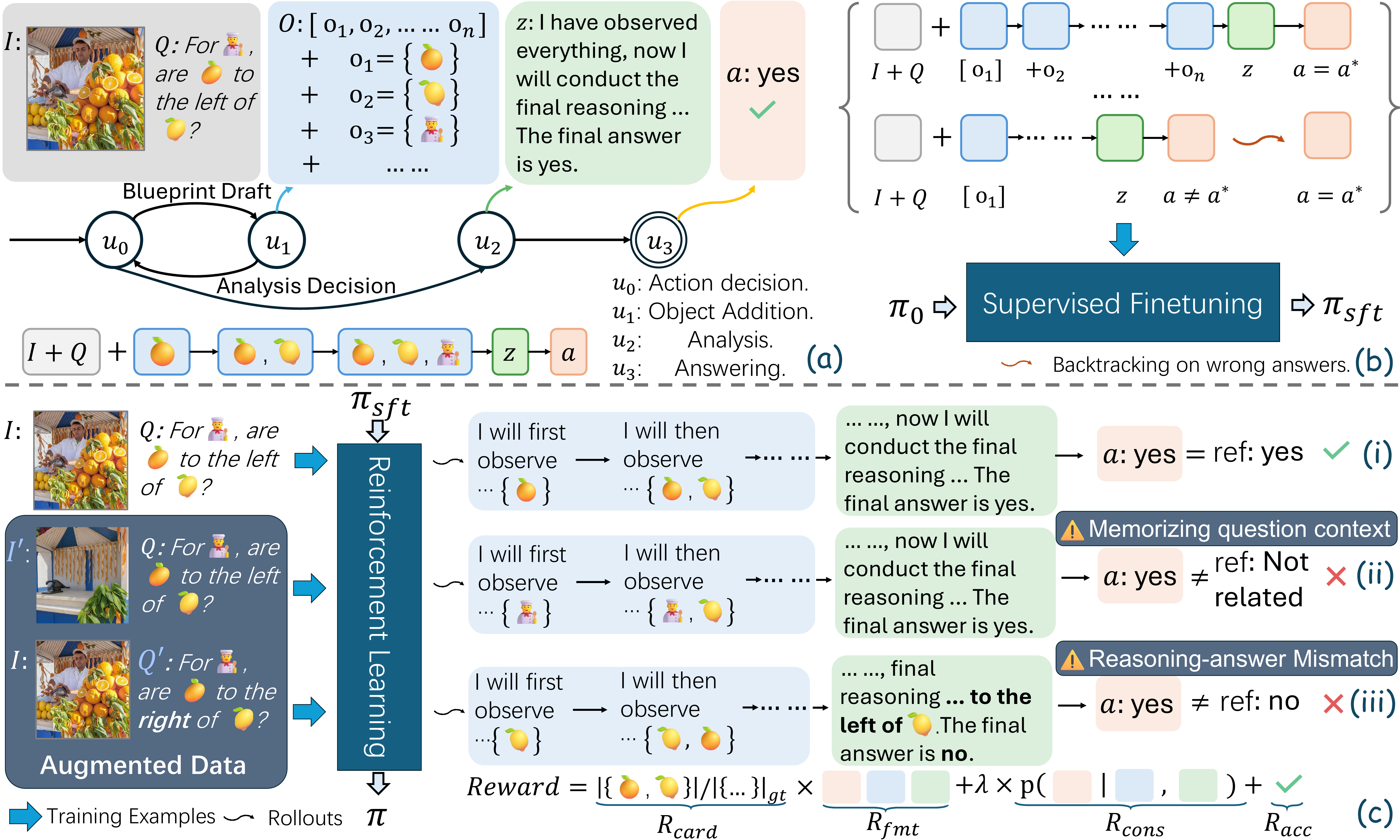}
    \caption{Approach overview. 
    \textbf{(a):} Construct \textbf{blueprint-embedded reasoning traces}.
    We prompt a strong teacher VLM to generate atomic reasoning steps, including adding objects to the blueprint, analyzing it, and producing the final answer. 
    These steps are then assembled into coherent traces via MCTS.
    \textbf{(b):} Perform supervised fine-tuning.
    The model is fine-tuned on the blueprint-embedded reasoning traces to elicit basic reasoning skills.
    \textbf{(c):} Perform reinforcement learning.
    The overall reward composed of two conventional ones (answer correctness and trace format) alongside two \textbf{blueprint-aware rewards}: object cardinality reward, which encourages including an appropriate number of objects in the blueprint, and causal consistency reward, which ensures final answers are grounded in intermediate reasoning.
    Moreover, we employ \textbf{anti-shortcut data augmentation}, perturbing images (example ii) or questions (example iii) to alter the original answer, preventing the model from relying on memorized visual or linguistic patterns.
    }
    \label{fig:general_pipeline}
\end{figure*}
\input{sec/3_method}

\input{sec/4_experiment}
\input{sec/5_conclusion}

\bibliographystyle{plainnat} 
\bibliography{main}

\appendix
\input{sec/X_suppl}

\end{document}

%% file: sec/0_abstract.tex
\begin{abstract}
Spatial reasoning---the ability to perceive and reason about relationships in space---advances vision–language models (VLMs) from visual perception toward spatial semantic understanding.
Existing approaches either revisit local image patches, improving fine-grained perception but weakening global spatial awareness, or mark isolated coordinates, which capture object locations but overlook their overall organization.
In this work, we integrate the cognitive concept of an object-centric blueprint into VLMs to enhance spatial reasoning.
Given an image and a question, the model first constructs a JSON-style blueprint that records the positions, sizes, and attributes of relevant objects, and then reasons over this structured representation to produce the final answer.
To achieve this, we introduce three key techniques: (1) blueprint-embedded reasoning traces for supervised fine-tuning to elicit basic reasoning skills; (2) blueprint-aware rewards in reinforcement learning to encourage the blueprint to include an appropriate number of objects and to align final answers with this causal reasoning; and (3) anti-shortcut data augmentation that applies targeted perturbations to images and questions, discouraging reliance on superficial visual or linguistic cues.
Experiments show that our method consistently outperforms existing VLMs and specialized spatial reasoning models.
\end{abstract}

%% file: sec/1_intro.tex
\section{Introduction}
\label{sec:intro}

\emph{Spatial reasoning} is a fundamental cognitive capability that reflects how humans perceive, understand, and interact with their surroundings.
It involves addressing typical questions such as \textit{Is A to the left of B?}, \textit{Can A fit into the gap around B?}, or \textit{Which turn should I take to reach the target?}~\citep{Kamath2023WhatsUp, ray2024sat, song2025robospatial, Yang2025ThinkingInSpace}.
Equipping \emph{vision–language models} (VLMs) with strong spatial reasoning abilities is essential---not only for enabling intelligent robotic perception and manipulation, but also as a key step toward developing more general and grounded forms of artificial intelligence.

Despite growing interest, current VLM-based approaches stochastically glimpse and infer, overlooking principled spatial layout modeling for rigorous reasoning (see lower half of Figure~\ref{fig:overall_illustration}).
One line of work~\citep{rose2023visual,zheng2025deepeyes,fu2025refocus} revisits local image patches through cropping or editing, called \emph{thinking with images}.
While these approaches enhance perception of fine-grained details, they weaken the awareness of a global spatial structure---the arrangement and mutual relations among objects.
Another line of work~\citep{Sarch2025ViGoRL,wu2025grounded,peng2023kosmos2,chen2023shikra,you2023ferret,zhang2024ferretv2} predicts or marks object positions using points or bounding boxes, called \emph{thinking with coordinates}.
Although effective for locating objects, they rely on scattered and isolated coordinates rather than organized entities that capture how objects relate in space.

Insights from cognitive science suggest that humans perceive and reason in space via a structured pathway: visual signals are first bound into individuated object files~\citep{kahneman1992reviewing,egly1994shifting}, which collectively form an object-centric blueprint encoding object-level layouts and relationships~\citep{tolman1948cognitive,okeefe1978hippocampus}. Reasoning then proceeds by scanning this blueprint to query spatial relations and produce a final judgment~\citep{kosslyn1973scanning,kosslyn1978visual}. Drawing parallels to prevailing VLMs~\citep{openai2025gpt5, qwenteam2025qwen2_5vl}, we identify analogous stages: large-scale pretraining provides general image perception, visual grounding via bounding boxes implements object-file binding and supports blueprint assembly, and chain-of-thought reasoning enables scanning the blueprint to reach final decisions. Together, these components form the foundation for linguistically replicating human-like spatial reasoning in VLMs.

Building on insights from cognitive science, we propose \textbf{thinking with blueprint}, chaining the pretrained capabilities of VLMs to emulate human cognitive process. Given an image and a question, our method first constructs a JSON-style blueprint that records the rough positions, sizes, and natural-language attributes of all objects relevant to the question, and then performs analysis over this structured representation to deliver the final answer (see the upper half of Figure~\ref{fig:overall_illustration}). This approach offers two key advantages. First, it separates the observation (constructing the blueprint) from the reflection (analyzing it), rather than entangling them as in prior approaches. Second, the blueprint provides a global and coherent spatial context for reasoning, helping the model move beyond fragmented or local cues.

To enable this, we first perform supervised fine-tuning (SFT) on VLMs to elicit basic reasoning skills using \textbf{blueprint-embedded reasoning traces}, and then apply reinforcement learning (RL) to further enhance this capability with \textbf{blueprint-aware rewards} and \textbf{anti-shortcut data augmentation} (see Figure~\ref{fig:general_pipeline}).
For \emph{blueprint-embedded traces}, since existing VLMs cannot directly produce such structured trajectories, we construct them through a stepwise collection pipeline. 
A strong teacher VLM is prompted to generate atomic reasoning steps, such as adding objects to the blueprint, analyzing it, or summarizing the final answer, which are then assembled into coherent and goal-directed traces via Monte Carlo Tree Search (MCTS).
For \emph{blueprint-aware rewards}, we introduce two forms of regulation: the first guides the blueprint to include an appropriate number of objects, penalizing incomplete blueprints that omit key information and capping rewards for those that include excessive irrelevant objects; while the second encourages consistency between the final answer and the blueprint-guided reasoning, ensuring that the model grounds its conclusions in the reasoning derived from the blueprint rather than relying on superficial correlations.
Finally, with \emph{anti-shortcut data augmentation}, we discourage shortcut learning by introducing perturbations to images or questions that disturb original answers.
This compels the model to reason through the constructed blueprint instead of relying on memorized visual or linguistic patterns.

We fine-tune Qwen2.5-VL~\citep{qwenteam2025qwen2_5vl} using the proposed techniques and evaluate it on representative spatial reasoning benchmarks.
Using a subset of SAT~\citep{ray2024sat} as training data, our method achieves a 35.9\% improvement over the base Qwen2.5-VL on this benchmark, and further yields gains of 4.3\%, 3.5\%, and 1.2\% on the out-of-distribution test sets from BLINK~\citep{Fu2024BLINK}, RoboSpatial~\citep{song2025robospatial} and VSR~\citep{Liu2023VSR}.
Moreover, our method surpasses proprietary models such as GPT-5-Thinking~\citep{openai2025gpt5} and specialized spatial reasoning models~\citep{Sarch2025ViGoRL, wu2025vilasr, yang2025machine}.
Our contributions are as follows:

\begin{itemize}
    \item We integrate the cognitive concept of object-centric blueprints into VLMs for spatial reasoning, enabling the model to first construct a JSON-style blueprint capturing object positions, sizes, and attributes, and then reason over this structured representation to produce answers.
    \item We introduce three key techniques into SFT\&RL pipeline to enable blueprint-based reasoning: SFT with blueprint-embedded traces to elicit basic skills, followed by RL with blueprint-aware rewards and anti-shortcut data augmentation for further improvement and generalization.
    \item Experiments demonstrate consistent gains over existing VLMs and specialized spatial reasoning models.
\end{itemize}

%% file: sec/2_related_work.tex
\section{Related Works}
\label{sec:related_work}

\noindent\textbf{Benchmarks.}
With growing interest in spatial reasoning, various datasets have emerged to improve or evaluate spatial awareness of VLMs.
These datasets span images~\citep{ray2024sat, Fu2024BLINK, Kamath2023WhatsUp, Liu2023VSR, cheng2024spatialrgpt}, videos~\citep{Yang2025ThinkingInSpace, Li2025STIBench, cheng2025vstar}, and 3D scenes~\citep{song2025robospatial, Azuma2022ScanQA, zhang2025open3d, cheng20253d}. 
Many of them provide only test sets~\citep{Fu2024BLINK, Kamath2023WhatsUp, Yang2025ThinkingInSpace, Li2025STIBench, zhang2025open3d, cheng2025vstar}. 
Among those with training sets, SAT \citep{ray2024sat} focuses on synthetic images, RoboSpatial \citep{song2025robospatial} generates questions from predefined geometric rules without verification, and VSR \citep{Liu2023VSR} and ScanQA \citep{Azuma2022ScanQA} cover only static relationships. 
A strong need remains for large-scale, real-world, and verified datasets for spatial reasoning.

\noindent\textbf{Adapting VLM Structures.}
Several works explore structural adaptations of VLMs to enhance spatial reasoning.
Some work alters attention maps to pinpoint the focus to target objects~\citep{Chen2025WhySpatialHard, Qi2025BeyondSemantics}.
Others introduce auxiliary encoders to incorporation spatial information, such as depth maps~\citep{chen2024spatialvlm, cheng2024spatialrgpt, cheng20253d} or 3D feature extractors~\citep{wu2025spatialmllmboostingmllmcapabilities, xu2024pointllm, xu2025pointllm}.
Additional approaches leverage spatial-temporal information from scene videos~\citep{yu2025videor1, ouyang2025spatialr1, yi2025stvlm, chen2025llavast, yuan2025videorefer}.

\noindent\textbf{Enabling Reasoning Capability.} 
Since Visual-CoT~\citep{rose2023visual}, enhancing VLMs' reasoning capability has become a prevalent approach across various tasks such as 2D wayfinding~\citep{wu2025vilasr, zhang2025latent}, GUI manipulation~\citep{Sarch2025ViGoRL}, document reasoning~\citep{luo2024layoutllm, liao2024doclayllm}, and robot manipulation~\citep{ye2025vla}.
Methods closely related to spatial reasoning can be roughly grouped into two categories.
The first, often called \emph{thinking with images}, augments models with explicit visual cues. 
These visual cues can be obtained by iteratively zooming or tiling regions of interest, highlighting key areas, calling external APIs such as OCR and chart parsers, or even outputting latent visual tokens~\citep{zheng2025deepeyes,wu2025vilasr,Yang2025Mirage, li2025lvr, zhang2025latent, wang2025pixel}.
The second, referred to as \emph{thinking with coordinates}, leverages explicit numeric coordinates to guide reasoning throughout the linguistic process~\citep{Sarch2025ViGoRL, rose2023visual, xu2025defacto}.
In contrast, we incorporate the cognitive concept of object-centric blueprints into VLMs to more closely emulate the human spatial reasoning process.

%% file: sec/3_method.tex
\section{Method}
\label{sec:method}
\subsection{Problem Formulation}
\label{subsec:problem}
Inspired by insights from cognitive science, we introduce the concept of \textbf{object-centric blueprints} into VLMs to enhance spatial reasoning (Figure~\ref{fig:overall_illustration}, upper half).
Given an image $I$ and a question $Q$, our method first constructs a JSON-style blueprint $O$ with $n$ objects relevant to the question, denoted as $O=[o_1;\dots;o_n]$ (where $;$ denotes concatenation).
The model then performs analysis $z$ over this structured representation to produce the final answer $a$, which is expected to match the ground-truth answer $a^*$.
Each object $o_i$ comprises a thought $s_i$ describing how the object is identified, an entity name $e_i$, a bounding box $b_i$ and a natural-language attribute $d_i$, i.e., $o_i=[s_i;e_i,b_i,d_i]$.
The complete reasoning trace $\tau$, which includes the blueprint $O$, the analysis $z$ and the final answer $a$, is thus represented as 
\begin{align}
\tau=[o_1;\dots;o_n; z; a].
\end{align}

Our goal is to train a vision-language model (VLM) $\pi_\theta$ that generates such a reasoning trace $\tau$ given the image $I$ and question $Q$, i.e., $\pi_\theta(\tau\mid I,Q)$.
Owing to the autoregressive nature of VLMs, this process can be factorized as
\begin{align}
    \pi_\theta(\tau\mid I,Q) = & \big(\prod_{i=1}^n\pi_\theta(o_i\mid I, Q)\big) \cdot \pi_\theta(z\mid I, Q, o_{\le n}) 
    \cdot \pi_\theta(a\mid I, Q, o_{\le n}, z).
\end{align}

\subsection{Approach Overview}
\label{subsec:overview}
To enable \textbf{thinking with blueprints}, we fine-tune the base VLM (Qwen2.5-VL in our implementation) using a two-stage recipe, i.e., supervised fine-tuning followed by reinforcement learning, with three key techniques: blueprint-embedded reasoning traces, consistency-preserving rewards, and anti-shortcut data augmentation.

\noindent\emph{\underline{1. Supervised Fine-tuning (SFT) Stage}} (Figure~\ref{fig:general_pipeline}(b)).
To elicit basic reasoning skills, we fine-tune the base VLM by minimizing the cross-entropy loss between the generated trace $\hat{\tau}$ and the ground-truth trace:
\begin{align}
    \mathcal{L}(\theta)=-\mathbb{E}_{(I,Q,\tau)\sim\mathcal{D}_{\text{SFT}}}\left[\frac{1}{T}\sum_{t=1}^T\log P_{\pi_\theta}\left(\hat{\tau_t}\mid I,Q\right)\right],
\end{align}
where $T$ is the length of the reasoning trace and $ P_{\pi_\theta}$ denotes the predicted probability of the $t$-th token in the trace.

The main challenge in this stage is that typical spatial reasoning datasets contain only image–question–answer triplets $(I, Q, a)$, without blueprint-guided reasoning traces $\tau$, which are required as supervision for SFT.
Moreover, existing VLMs cannot directly produce such blueprint-based traces.
To address this, we construct \textbf{blueprint-embedded reasoning traces} through a stepwise collection pipeline.
Specifically, we prompt a strong teacher VLM to generate atomic reasoning steps and then assemble them into coherent and goal-directed traces using Monte Carlo Tree Search (MCTS).
Details are provided in \textbf{Section~\ref{subsec:blueprint_construction}}.

\noindent\emph{\underline{2. Reinforcement Learning (RL) Stage}} (Figure~\ref{fig:general_pipeline}(c)).
To further strengthen the model’s reasoning capability, we apply RL to directly optimize the reasoning behavior sampled from the base model by maximizing the expected reward over reasoning traces $\tau$:
\begin{align}
\max_\theta\mathbb{E}_{(I,Q,a)\sim\mathcal{D}_{\text{RL}}}\left[\mathbb{E}_{\tau\sim\pi_\theta}\left[R(\tau)\right]\right].
\end{align}
Here we adopt Group Relative Policy Optimization (GRPO)~\citep{shao2024deepseekmath} in our implementation.

The success of this stage hinges on effective reward design.
While prior spatial reasoning methods only reward answer correctness and trace format~\citep{Sarch2025ViGoRL}, the introduction of blueprints makes \textbf{blueprint-aware rewards} essential.
First, if the blueprint includes too few objects, the subsequent analysis may be based on incomplete information; conversely, if it includes too many, irrelevant objects may dominate and mislead the reasoning process.
To address this, we introduce a reward that encourages the blueprint to include an appropriate number of objects, penalizing insufficient object cardinality and capping rewards for excessive ones.
Second, even when the blueprint is relevant and the reasoning coherent, the model may still generate a final answer that contradicts its own reasoning.
Inspired by recent work~\citep{yu2025rlpr}, we introduce another reward that enforces consistency between the final answer and the blueprint-guided reasoning trace.
Details are provided in \textbf{Section~\ref{subsec:reward}}.

Moreover, we observe shortcut behaviors~\citep{ye2025prism,xia2025visionaryr1} during the RL stage, where the model tends to take easier but less generalizable paths to solve the question.
In particular, it sometimes relies on memorizing visual or linguistic patterns instead of constructing proper blueprints and performing coherent reasoning.
As illustrated in Figure~\ref{fig:general_pipeline}(c), example (ii) shows a case where the model derives the final answer by memorizing the question while ignoring image changes, whereas example (iii) shows the opposite behavior.
To mitigate such shortcuts, we introduce \textbf{anti-shortcut data augmentation}, which perturbs the images or questions in ways that distort the original answer, thereby encouraging the model to reason rather than memorize.
Details are provided in \textbf{Section~\ref{subsec:augmentation}}.

\subsection{Constructing Blueprint-Embedded Traces}
\label{subsec:blueprint_construction}

Inspired by prior work~\citep{Sarch2025ViGoRL}, we employ Monte Carlo Tree Search (MCTS) to generate blueprint-embedded reasoning traces, which are used as supervision during the SFT stage (Figure~\ref{fig:general_pipeline}(a)).
In our formulation, each node in the search tree represents an atomic reasoning step---either (1) adding an object $o_i$ to the blueprint, (2) performing an analysis $z$ over the constructed blueprint, or (3) summarizing the final answer $a$. 
A strong VLM is prompted at each step to propose the next atomic action.
The search begins from the root node initialized by the input image–question pair $(I, Q)$. Internal nodes correspond to incremental blueprint construction by adding objects $o_i$, while leaf nodes terminate with the analysis $z$ and final answer $a$.
Each rollout is evaluated by the answer correctness, and the score is back-propagated through the tree to guide exploration toward more promising reasoning paths.
Finally, we linearize the root-to-leaf paths into blueprint-embedded reasoning traces. 
We retain both the traces that lead to correct answers and those that trigger backtracking to correct initial failed rollouts~\citep{Sarch2025ViGoRL}.

\subsection{Blueprint-Aware Rewards}
\label{subsec:reward}
Regarding the rewards used in the RL stage, we adopt two commonly used rewards from prior work (i.e., answer correctness and trace format)~\citep{Sarch2025ViGoRL} and introduce two additional rewards that are crucial for improving the performance of blueprint-based reasoning (i.e., object cardinality and causal consistency).
The overall reward $R$ is defined as:
\begin{align}
    R = R_{\text{acc}} + R_{\text{fmt}}\cdot R_{\text{card}} +R_{\text{cons}}.
\end{align}
Specifically, (1) \emph{answer correctness} $R_{\text{acc}}$ evaluates whether the generated answer $a$ matches the ground-truth answer $a^*$; (2) \emph{trace format} $R_{\text{fmt}}$ verifies that the reasoning traces use the correct \texttt{<think>} and \texttt{<answer>} tags and that the blueprint can be properly parsed as JSON; (3) \emph{object cardinality} $R_{\text{card}}$ encourages the blueprint to include an appropriate number of objects, penalizing too few and capping rewards for excessive ones; and (4) \emph{causal consistency} $R_{\text{cons}}$, encourages the model to ground its final answer in the reasoning derived from the blueprint.
We use the product $R_{\text{card}} \cdot R_{\text{fmt}}$ rather than their sum because $R_{\text{fmt}}$ tends to saturate early in training. 
Using it as a multiplier allows the model to focus more on improving $R_{\text{card}}$ once $R_{\text{fmt}}$ stabilizes, while still maintaining valid formatting.
In the following, we detail the two newly introduced rewards.

\noindent\textbf{Object Cardinality Reward.}
We use the number of objects mentioned in the question or answer, denoted as $K$, as a reference to evaluate whether the blueprint includes an appropriate number of objects.
Specifically, $K$ is precomputed either by extracting it directly from the answer (e.g., $K=4$ for “Q: How many chairs are there? A: 4”) or by counting the number of objects explicitly mentioned in the question (e.g., $K=2$ for “Q: What is the spatial relationship between the sofa and the table?”).
Let $|O|$ denote the number of objects in the generated blueprint. When $|O| \le \lambda K$, the model is rewarded to encourage exploration of a sufficient number of objects. When $|O| > \lambda K$, the reward is capped to prevent the model from including excessive irrelevant objects.
Here, $\lambda$ is an integer controlling the tolerance and is set to 2 in our implementation.
Formally, the object cardinality reward is defined as
\begin{align}
R_{\text{card}} = \min\{|O| / K, \lambda\}.
\end{align}
When counting objects in the blueprint, we only consider those with distinct positions and sizes, determined by an IoU threshold of $\le 0.3$ between bounding boxes.

\noindent\textbf{Causal Consistency Reward.}
We draw inspiration from prior work RLPR~\citep{chen2025reasoningfaithfulness}.
While RLPR was originally proposed to replace verifier-based rewards with probability-based rewards, its underlying principle, that the model’s intrinsic probability of generating the correct answer reflects how faithfully its reasoning supports that answer, aligns well with our objective of encouraging the model to ground its final answer in the reasoning process.
Thus, we adopt RLPR’s probability-based formulation and define the causal consistency reward as the difference between the average teacher-forcing logits of the answer tokens $a^*$ conditioned on the full reasoning context $[I; Q; O; z]$ and those conditioned only on the input pair $[I; Q]$:
\begin{align}
    R_{\text{cons}}=\frac{1}{T_a}\sum_{t=1}^{T_a}\big(P_{\pi_\theta}(a_t\mid a_{<t},I,Q,O,z) - P_{\pi_0}(a_t\mid a_{<t},I,Q)\big),
\end{align}
where $T_a$ is the answer length.

\subsection{Anti-Shortcut Data Augmentation}
\label{subsec:augmentation}

As illustrated in examples (ii) and (iii) of Figure~\ref{fig:general_pipeline}(c), the model may derive the final answer by memorizing the question (or image) while ignoring changes in the image (or question). 
In other words, it tends to follow easier but less generalizable paths. 
To mitigate such shortcut behaviors, we augment the RL-stage data by perturbing the image or question and modifying the original answer accordingly.

For a given image $I$, we generate a perturbed version $I^\prime$ as follows. 
First, we prompt an LLM (GPT-4o in practice) to identify the objects mentioned in the question. 
Next, we use these object names to construct editing instructions and employ an image-editing model (Flux-Kontext~\citep{labs2025flux} in practice) to remove the specified objects, producing $I^\prime$. The corresponding altered answer $a^\prime$ is set to ``0'' for counting problems and to ``question and image do not match'' for all other types of question.
For a given question $Q$, we generate a perturbed version $Q^\prime$ as follows. 
We define a prompt template to instruct the LLM to identify spatial predicates related to locations, directions, and other actions. 
The LLM then selects predicates that are most likely to invert the original answer.
The corresponding altered answer $a^\prime$ is set to the opposite of the original answer. 
For example, in Figure~\ref{fig:general_pipeline}(c), example (iii) is an augmentation of example (i), where the spatial predicate ``left'' is changed to ``right,'' and the answer is altered from ``yes'' to ``no''.
Finally, to further reduce potential noise, we use an LLM to filter out augmented examples $[I^\prime, Q, a^\prime]$ and $[I, Q^\prime, a^\prime]$ where $[I^\prime, Q]$ or $[I, Q^\prime]$ do not actually lead to $a^\prime$.

%% file: sec/4_experiment.tex
\begin{table*}[]
\centering
{\setlength{\tabcolsep}{1pt}
\begin{tabular}{lclccccc}
\toprule[1.5pt]
          Method   & Model Size & Reasoning Mode & SAT val & SAT test & Blink & Robospatial & VSR\\
& & & \textit{(iid)} & \textit{(ood)} & \textit{(ood)} & \textit{(ood)} & \textit{(ood)} \\            
\midrule[1.2pt]
\rowcolor{gray!20}
\multicolumn{8}{@{}l}{\textit{Proprietary Models}} \\
gpt-4o   & -      & No &  57.2   & 51.5        & 59.2 & 60.1 & 78.7 \\
gpt-4o & -  & Naive CoT & 57.7 & 63.3 & \underline{59.0} & 63.6 & 82.2 \\
gpt-5-Thinking & -  & Naive CoT & 58.3 & 72.7 & 56.3 & 65.4 & \underline{84.2} \\
\midrule
\rowcolor{gray!20}
\multicolumn{8}{@{}l}{\textit{Open-sourced Models}} \\
Qwen2.5-VL &  7B  & No & 56.8    & 63.3     & 56.4  & 66.7 & 83.6 \\
Qwen2.5-VL &  7B  & Naive CoT  & 52.6    & 54.7     & 53.8  & 66.3 & 82.2 \\
Robix & 7B  & Naive CoT &  -   & 71.1     & -$^{*}$  & - & 83.3 \\
Robix & 32B   & Naive CoT & -    & \underline{79.6}      & -$^{*}$  & -  & 83.7 \\
\midrule
\rowcolor{gray!20}
\multicolumn{8}{@{}l}{\textit{Specialized Spatial Reasoning Methods}} \\
VigoRL & 7B      & Thinking with Coordinates & \underline{67.5}    & 57.3     & 56.1  & \underline{67.9} & 82.7 \\
ViLASR & 7B        & Thinking with Images  & 57.7    &  60.7    &  53.3 & 48.5 & 76.5\\
Mirage & 7B      & Thinking with (Latent) Images  & -$^{*}$    & 72.0     & -$^{*}$  & - & - \\
\midrule
\rowcolor{gray!20}
\multicolumn{8}{@{}l}{\textit{Our Method}} \\
Ours & 7B          & Thinking with Blueprint & \textbf{92.7}    & \textbf{79.7}     & \textbf{60.7}  & \textbf{70.2} & \textbf{84.8} \\
\multicolumn{3}{@{}l}{\hspace{0.1em}\textit{Gain over Qwen2.5-VL}}  & \textit{+35.9} & \textit{+16.4} & \textit{+4.3} & \textit{+3.5} & \textit{+1.2}\\
\bottomrule[1.5pt]
\end{tabular}
}
\caption{
Quantitative results. 
The best is in \textbf{bold} and the second best is \underline{underlined}.
SAT val follows a similar distribution as the training set (denoted as \emph{iid}), while SAT-test, Blink, Robospatial, and VSR differ from the training distribution (denoted as \emph{ood}).
``-'' indicates that results are unavailable or the method is difficult to reproduce on the benchmark.
``-$^{*}$'' indicates results obtained under training or evaluation settings different from ours (see Appendix for details).
Our method achieves the best performance across both iid and ood settings, outperforms substantially larger models (e.g., GPT series and Robix-32B) and surpasses other specialized spatial-reasoning approaches.
}
\label{tab:quantitaive_res_1}
\end{table*}

\section{Experiment}
\label{sec:experiment}

\subsection{Experiment Setup}
\textbf{Training Datasets.} 
We use the SAT training split~\citep{Ray2025SAT}, which consists of synthetic indoor scene images, following prior work~\citep{Sarch2025ViGoRL}.
For SFT, we collect 65k blueprint-embedded reasoning traces by running MCTS rollouts on 1.5k samples with GPT-4o, using the method described in Section~\ref{subsec:blueprint_construction}.
For RL, we sample 32k question–answer pairs from the SAT training split, generating one augmented variant for each using the procedure in Section~\ref{subsec:augmentation}.

\noindent\textbf{Training Configuration.} 
For SFT, we start from Qwen2.5-VL Instuct 7B \citep{qwenteam2025qwen2_5vl} and finetune it with LlamaFactory \citep{zheng-etal-2024-llamafactory}, using AdamW optimizer with learning rate 1e-6, global batch size 32 and gradient accumulation step 4 over 1500 steps.
We then perform RL for 500 steps with learning rate 1e-6 using 8 B200 Nvidia GPUs . 
For each step, 768 samples are taken, with each sample 8 rollouts. The global batch size for each update is 64.

\noindent\textbf{Evaluation Benchmarks.}
We evaluate our method on five representative spatial reasoning benchmarks.
The SAT validation set (SAT val)~\citep{Ray2025SAT}, which shares a similar distribution with our training data, is marked as \emph{iid} in Table~\ref{tab:quantitaive_res_1}.
The SAT test set (SAT test), BLINK~\citep{Fu2024BLINK}, Robospatial~\citep{song2025robospatial}, and VSR~\citep{Liu2023VSR} differ from the training distribution, either containing real-world images or involving unseen question types (e.g., visual correspondence or compatibility), and are therefore labeled \emph{ood} in Table~\ref{tab:quantitaive_res_1}.
For BLINK and Robospatial, we use only the subsets directly related to spatial reasoning (see Appendix for the full list).
Our model is trained solely on the SAT training split and evaluated on all benchmarks without any task-specific fine-tuning.

\noindent\textbf{Baseline Methods.}
We compare against proprietary VLMs, open-sourced VLMs, and prior spatial-reasoning methods. 
For \emph{proprietary models}, we evaluate GPT-4o \citep{hurst2024gpt4o} (with/without CoT) and reasoning-native GPT-5-Thinking \citep{openai2025gpt5} (medium thinking level) for generic spatial reasoning. 
For \emph{open-sourced models}, we include the generalist Qwen2.5-VL \citep{qwenteam2025qwen2_5vl} (the base VLM used in our approach) in both direct-answer and CoT modes, as well as the specialist Robix \citep{huang2025robix} across multiple parameter scales. 
For \emph{specialized spatial reasoning methods}, we cover models with thinking-with-coordinates mechanism (ViGoRL \citep{Sarch2025ViGoRL}) and thinking-with-image mechanism (ViLASR \citep{wu2025vilasr} with cropping/zooming/box/line tools).
We also report Mirage \citep{yang2025machine}, which performs reasoning over image latents.

\begin{figure*}[t]
    \centering
    \includegraphics[width=\linewidth]{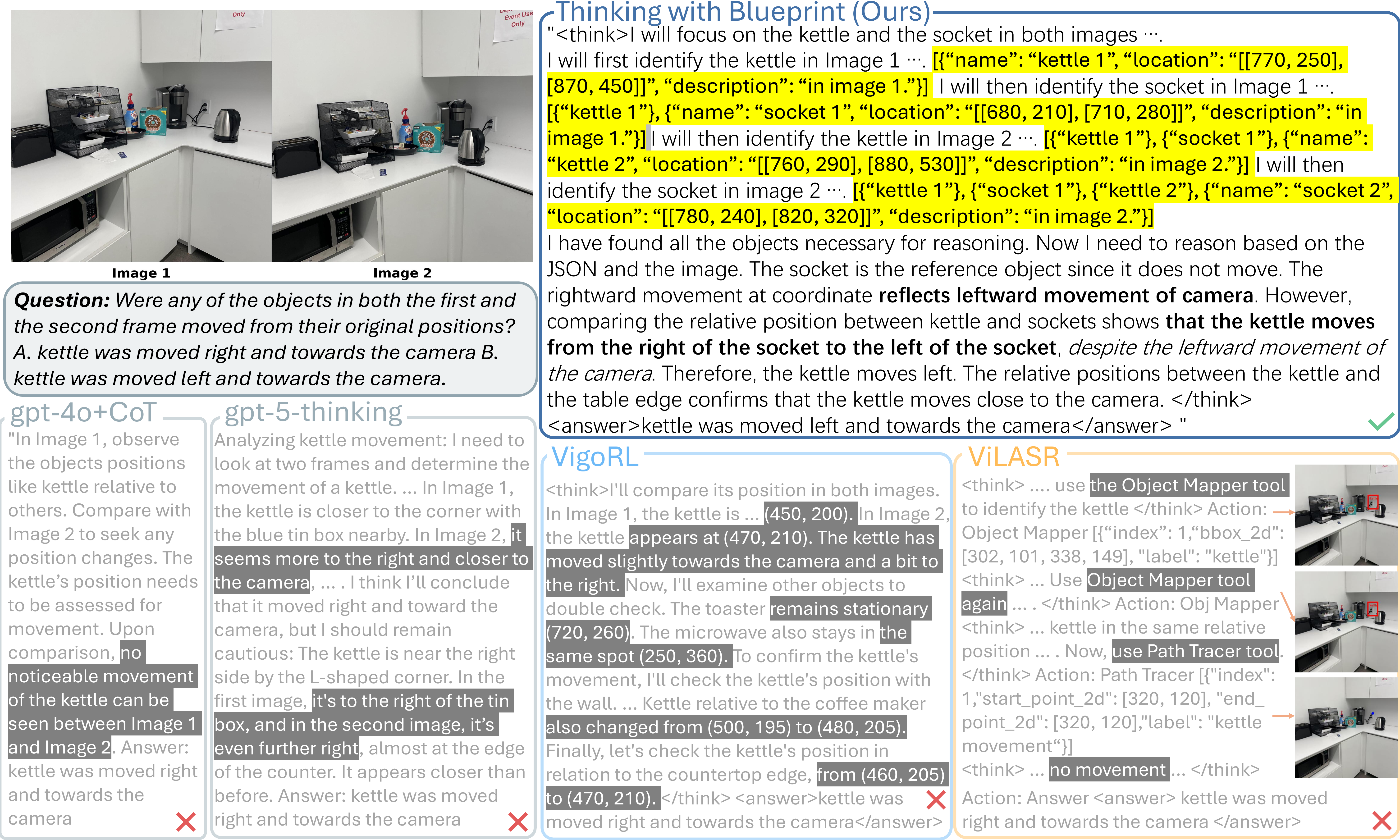}
    \caption{
    Qualitative results.
    Between Image 1 and Image 2, both the camera position and the kettle position change, and the question asks how the kettle moves.
    With the blueprint serving as a well-structured representation, our model identifies the socket as a fixed reference and correctly infers the kettle’s movement.
    Other methods either miss the movement entirely or fail to account for the camera motion.
    }
    \label{fig:qualitative_res}
\end{figure*}

\subsection{Quantitative Results}

Table \ref{tab:quantitaive_res_1} reports the quantitative results.
On the \emph{iid} benchmark (SAT-val), our method surpasses all baselines, yielding a 35.9\% improvement over the base Qwen2.5-VL and roughly a 15\% gain over the strongest specialized spatial-reasoning method (ViGoRL).
On the \emph{ood} benchmarks (SAT-test, Blink, Robospatial, and VSR), despite no task-specific finetuning, our method still outperforms all baselines, including substantially larger models such as the GPT series and Robix-32B.
Below we analyze key insights and observations from these results. 

\noindent\textbf{Discovery 1: Reasoning is not always effective.} 
For Qwen2.5-VL, its native CoT version underperforms the non-reasoning version across all benchmarks.
Specialized spatial-reasoning methods such as ViGoRL and ViLASR, despite being finetuned from the Qwen2.5-VL, often perform worse than the base model on several benchmarks.
Even for very large VLMs like GPT-4o, the native CoT version yields limited gain on SAT val and degrade performance on Blink.
These observations highlight that an effective reasoning strategy is crucial for achieving gains.

\noindent\textbf{Discovery 2: Organization of perceived content matters.}
ViGoRL, ViLASR, and our method all leverage numerical spatial cues (e.g., bounding boxes) during reasoning.
However, our method outperforms ViGoRL and ViLASR across all benchmarks.
The key distinction is that we introduce a blueprint to systematically organize all observations before analysis, while ViGoRL and ViLASR interleave scattered observation and reasoning.
(see qualitative examples in Figures \ref{fig:overall_illustration} and \ref{fig:qualitative_res}.)
This suggests that a well-structured blueprint is crucial for enhancing performance.

\begin{figure*}[t]
    \centering
    \includegraphics[width=\linewidth]{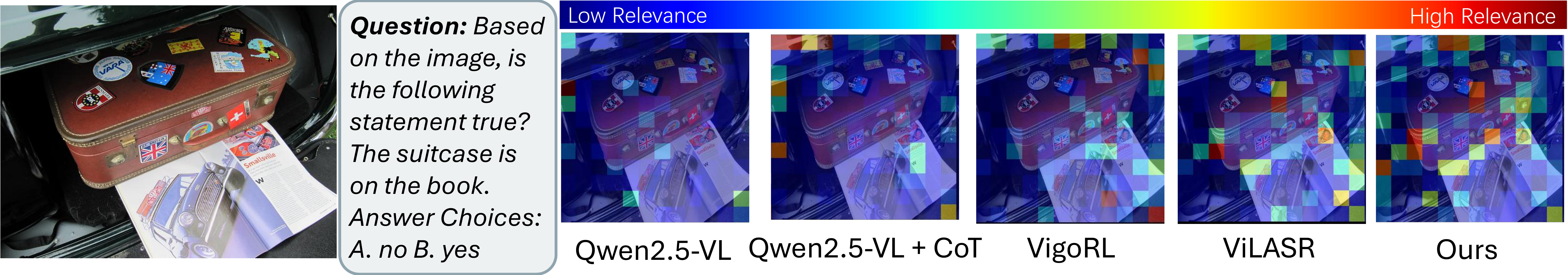}
    \caption{
    Visualization of attention maps follows prior work~\citep{Chen2025WhySpatialHard}.
    In our method, high-relevance image patches cluster tightly around the true region of interest, whereas in other methods they tend to scatter.
    }
    \label{fig:attn_map}
\end{figure*}

\noindent\textbf{Discovery 3: Other factors also contribute to failures.}
Beyond the reasoning strategy, additional factors can also affect performance. 
For example, in ViLASR, some errors arise from tool-call failures (e.g., array out-of-bounds, segmentation faults), which collapse the entire reasoning trace. 
In GPT-5-Thinking, failures can occur when reasoning exceeds the model’s maximum token limit.

\subsection{Qualitative Results}

\noindent\textbf{Visualization of Reasoning Traces.} 
Figure \ref{fig:qualitative_res} presents a qualitative comparison.
In this example, between Image 1 and Image 2, both the camera position and the kettle position change, and the question asks how the kettle has moved.
GPT-4o and ViLASR both miss the kettle’s movement.
For ViLASR, this failure originates from an incorrect kettle localization (red bounding box).
GPT-5-Thinking and ViGoRL detect the kettle’s movement but fail to account for the camera motion, leading to incorrect conclusions.
In our method, with the blueprint as a structured and comprehensive scene representation, the model identifies the wall socket as a fixed reference in world coordinates and correctly uses it to infer the kettle’s actual movement.

\begin{wraptable}{r}{3.5in}
\centering
\begin{tabular}{lccc}
\toprule[1.5pt]
\multicolumn{2}{@{}r}{SAT val}    & SAT test & Robospatial\\
\multicolumn{2}{@{}r}{\emph{(iid)}}    & \emph{(ood)} & \emph{(ood)}\\
\midrule[1.2pt]
\textbf{(A)} Ours                 & 92.7    & 79.7    & 70.2 \\
\textbf{(B)} $-$ Data Augmentation & 91.4    & 68.3    & 62.9 \\
\textbf{(C)} $-$ $R_\text{card}\; \&\; R_\text{cons}$            & 83.7    & 63.6  & 59.8  \\
\textbf{(D)} Vanilla GRPO          & 58.3    & 56.3  & 64.7 \\
\textbf{(E)} SFT only         & 70.1    & 68.7  & 53.5 \\
\bottomrule[1.5pt]
\end{tabular}
\caption{
Ablation study.
\textbf{(A)} Full version of our approach.
\textbf{(B)} Without anti-shortcut data augmentation.
\textbf{(C)} Without blueprint-aware rewards.
\textbf{(D)} Vanilla GRPO: no SFT, no blueprint, and none of the blueprint-related techniques.
\textbf{(E)} Without the RL stage, while retaining SFT on blueprint-embedded traces.
}
\label{tab:ablation}
\end{wraptable}

\noindent\textbf{Visualization of Attention Maps.}
To study how different methods shape visual focus, we visualize attention maps following prior work~\citep{Chen2025WhySpatialHard}.
We project the normalized attention between the first answer token and all image patches back onto the image.
Figure~\ref{fig:attn_map} shows the results, where cooler colors indicate lower relevance and warmer colors indicate higher relevance.
Reasoning-based methods, i.e., Qwen2.5-VL+CoT, ViGoRL, ViLASR, and ours, generally sharpen attention on specific image patches.
However, in Qwen2.5-VL+CoT, ViGoRL, and ViLASR, the high-attention patches tend to be scattered, whereas in our method, they concentrate around the true region of interest.
This tighter and more semantically aligned focus may help explain our method’s improved performance.

\subsection{Ablation Study}
Table~\ref{tab:ablation} summarizes the ablation results.
First, removing anti-shortcut data augmentation (row \textbf{(B)}) causes modest degradation on the \emph{iid} benchmark (SAT-val) but substantial drops on \emph{ood} benchmarks (SAT-test and Robospatial), highlighting its key role in improving generalization.
Second, removing blueprint-aware rewards (row \textbf{(C)}) leads to clear performance declines across all benchmarks, underscoring their importance.
Third, vanilla GRPO (row \textbf{(D)}) performs significantly worse than our method, demonstrating the effectiveness of the blueprint-based spatial reasoning workflow as a strong foundation for endowing spatial awareness in VLMs.
Finally, removing the RL stage (row \textbf{(E)}) also yields performance degradation across all benchmarks.

\begin{wrapfigure}{r}{3.28125in}
    \vspace{-30pt}
    \centering
    \includegraphics[width=3.28125in]{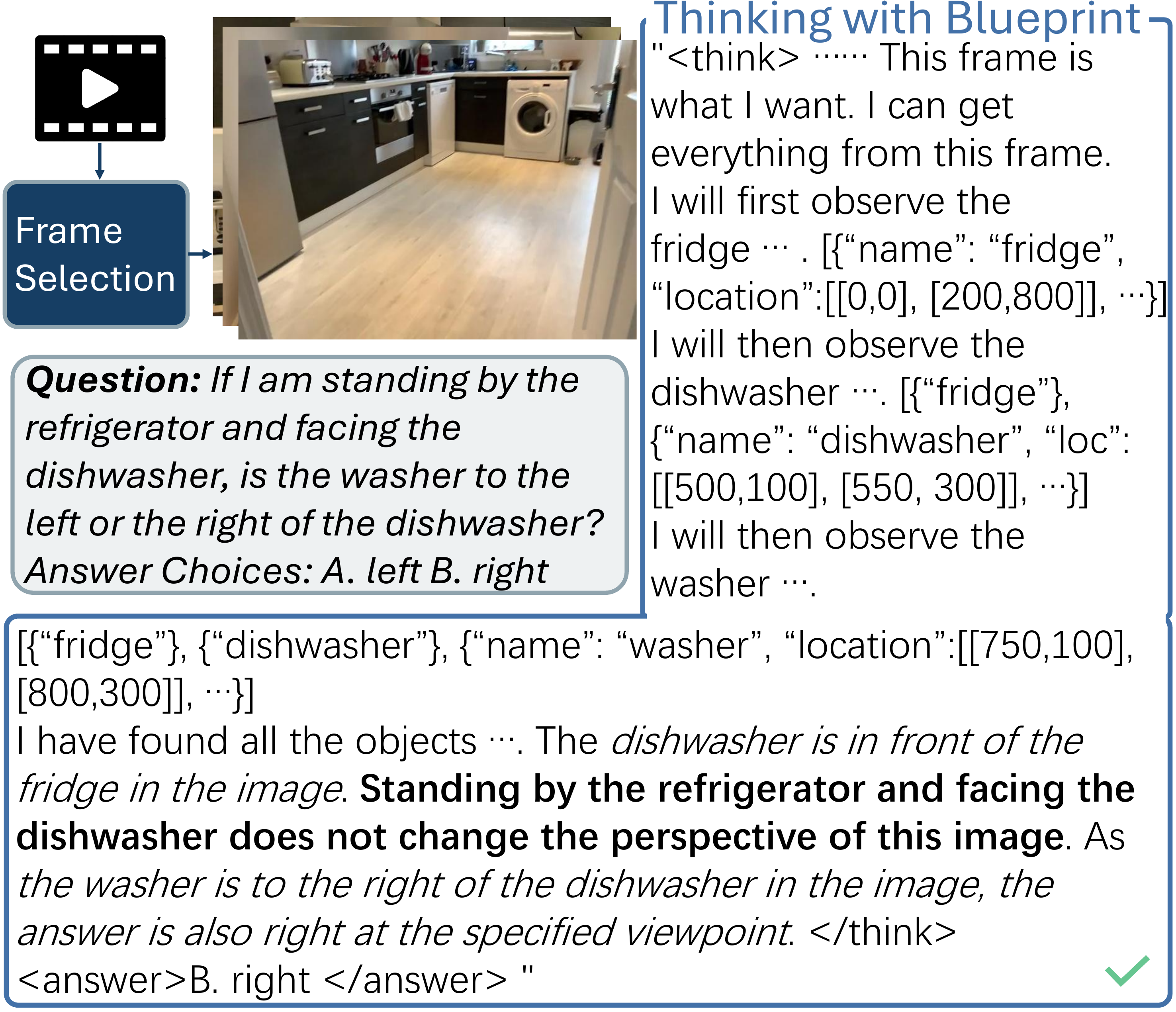}
    \caption{
    Example of extending our method to video-based spatial reasoning using a frame-selection module.
    }
    \label{fig:vsi}
\end{wrapfigure}

\subsection{Potential Extension}
While our method is trained only on images, we find that it can be applied to videos without additional retraining.
Figure~\ref{fig:vsi} provides an example.
The key idea is to introduce a lightweight frame-selection module.
Following prior work~\citep{zhang2025q}, we order the input frames of the video based on CLIP similarity between the frames and the question, and select the top four important frames as the input.
We observe that our method performs well \textit{as long as there exists the frames with all the mentioned objects clearly visible}.
This suggests a promising direction: by improving the frame-selection module, it is possible to adapt image-trained spatial reasoning models to videos.
Additional details are provided in the Appendix.

%% file: sec/5_conclusion.tex
\section{Conclusion}
\label{sec:conclusion}
Building on insights from cognitive science, we propose thinking with blueprint to enhance VLMs' spatial reasoning. 
Given an image and a question, our method constructs a JSON-style blueprint recording object positions, sizes, and attributes, and then analyzes over it to produce the answer. 
To achieve this, we perform SFT using blueprint-embedded reasoning traces to elicit basic reasoning skills, followed by RL with blueprint-aware rewards and anti-shortcut data augmentation for further improvement and generalization. 
In the future, we plan to extend this approach to video- and 3D-based spatial reasoning and to empower downstream tasks, such as robotic manipulation, with spatially enhanced VLMs.

\noindent\textbf{Limitations.}
(1) Our training relies on SAT, which contains synthetic scene images; a more diverse dataset could further improve performance.
(2) Larger VLMs (e.g., 32B) have not been explored due to limited computational resources.

%% file: sec/X_suppl.tex
\clearpage
\setcounter{page}{1}

\section{Details of the Reinforcement Learning Stage}
\label{sec:rl_detail}
Group Relative Policy Optimization (GRPO) is adopted as the advantage estimator of the reinforcement learning process, which improves the stability of policy training on long-form trajectories by using group-wise normalized advantages and applying a clipped, token-level PPO-style objective.

Given a group of $G$ trajectories $\mathcal{O} = {\tau^{(i)}}_{i=1}^{G}$ conditioned on an input $x$, each trajectory $\tau^{(i)}$ receives a scalar reward $r^{(i)} = R(\tau^{(i)})$. GRPO then forms a centered advantage $\hat{A}^{(i)} = r^{(i)} - \bar{R}$, where the baseline $\bar{R} = \frac{1}{G} \sum_i r^{(i)}$ is the average reward over the group.

Let $\tau^{(i)}_t$ denote the $t$-th token of trajectory $\tau^{(i)}$. The GRPO objective is the following clipped surrogate loss:
\begin{equation}
\begin{aligned}
\mathcal{L}_{\text{GRPO}}(\theta)
&=
-\frac{1}{G}
\sum_{i=1}^{G}
\frac{1}{|\tau^{(i)}|}
\sum_{t}
\min \Bigl[
  \rho^{(i)}_t \hat{A}^{(i)},
  \text{clip}(\rho^{(i)}_t, 1{-}\varepsilon, 1{+}\varepsilon)\hat{A}^{(i)}
\Bigr] + \beta\,\mathrm{KL}[\pi_\theta \Vert \pi_{\text{ref}}].
\end{aligned}
\end{equation}
where $\rho^{(i)}_t = \frac{\pi_\theta(\tau^{(i)}_t \mid \tau^{(i)}_{<t}, x)}{\pi_{\text{old}}(\tau^{(i)}_t \mid \tau^{(i)}_{<t}, x)}$ is the importance sampling ratio, $\varepsilon = 0.2$ is the clipping threshold, and $\beta$ controls the strength of the KL regularization toward the reference policy $\pi_{\text{ref}}$. 
For the advantage estimator $A$, all the advantages whose rewards falls inside the standard deviation range around the mean value are zeroed out, which is formally expressed as follows. 
\begin{equation}
\begin{aligned}
\tilde{R}_i = & \frac{ R(y_i) - \mathrm{mean}\!\left(R_G\right) }
     { \mathrm{std}\!\left(R_G\right) } \cdot \mathds{1} (|R(y_i) - \mathrm{mean}\!\left(R_G\right)| \geq \mathrm{std}\!\left(R_G)\right)), \\
\hat{A}_{i,t} = & \tilde{R}_i \; (\forall\, t \in \{1,\ldots,|y_i|\}), \\ R_G = & \{R(y_1),\ldots,R(y_G)\},\ R(y_i) \in R_G.    
\end{aligned}
\end{equation}

Such formulation is particularly effective for stabilizing optimization in long-horizon, multimodal reasoning tasks.

\section{Details of Data Augementation Pipeline}
\label{sec:data_aug_detail}
Here we detail the process of our data augmentation pipeline. For SAT training set, we let the VLM (in practice gpt-4o) to decide whether to edit the image $I$ and augment it into $I'$ and remain the question $Q$ unchanged, or to perturb the question $Q$ and obtain $Q'$ which leads to the reverse of the original answer $a$ and remain the image $I$ unchanged.

For augmenting the image $I$ and augment it into $I'$, the process is as follows. At first, both $Q$ and $I$ are sent to a VLM (in practice GPT-4o) to detect the subset of the objects $O$ in the image $I$ that relates to the question $Q$. Then for each object $o_k \in O$, we let the VLM write an object removal prompt $t_k$. Finally all {$t_k$} are sent to an image editor one by one, until all the concerning objects in the image are removed.

For augmenting the question $Q$ into $Q'$, the process is as follows. We combine both $Q$ and $I$, along with a prompt template $T$ which contains a set of examples of reversing the question, and send them to a VLM (in practice GPT-4o) to let the VLM decide the modified question $Q'$. The full prompt template about the process of editing images and questions is shown in Table \ref{tab:editing_prompts_1} and Table \ref{tab:editing_prompts_2}.

Finally, all the pairs of $(I, Q', a')$ and $(I', Q, a')$ are sent to the VLM again so as to assure the new answer adheres to the image-question-answer triplet, and the augmented question / answer is plausible.

\begin{table*}[t]

\centering
\begin{tabularx}{\textwidth}{>{\ttfamily\footnotesize}X }
\toprule
\textbf{Prompt template for testing proprietary models.} \\
prompt\_gpt4o = 
    "You are a careful visual QA assistant. Given an image and a multiple-choice question, answer the EXACT TEXT of the correct answer choice wrapped by <answer> and </answer>, verbatim from the options. Do not add letters (A/B), punctuation, or extra words in your answer. " \\
prompt\_gpt4o\_cot = "You are a careful visual QA assistant. Given an image and a multiple-choice question, answer the EXACT TEXT of the correct answer choice wrapped by <answer> and </answer>, verbatim from the options. Do not add letters (A/B), punctuation, or extra words in your answer. " \\ Think step by step before answering. Put your thinking prrocess between <think> and </think> tags." \\
prompt\_gpt5\_thinking = "You are a careful visual QA assistant. Given an image and a multiple-choice question, answer the EXACT TEXT of the correct answer choice wrapped by <answer> and </answer>, verbatim from the options. Do not add letters (A/B), punctuation, or extra words in your answer. " \\
\bottomrule
\end{tabularx}
\caption{Prompt template for testing proprietary models like gpt-4o, gpt-4o with CoT and gpt-5-thinking.}
\label{tab:prompts_gpt}
\end{table*}

\section{More Details about Experiment Settings}
\label{sec:more_exp_details}
\subsection{Details of data selection in the training set.}
SAT is used as training set throughout the reinforcement learning stage. In practice, to make a fair comparison with VigoRL, we conduct an even sampling with the same amount of the total QA pairs used. In particular, we randomly sample 24373 samples from SAT static, 3071 from action consequence, 2313 from action sequence, 1290 from object movement, 1233 from goal aim, and 442 from perspective.

\subsection{Full list of subcategories used in Blink.}
For testing Blink dataset, we use 11 of the 14 categories. The full list of categories used is: Visual Correspondence, Jigsaw, Spatial Relation, Semantic Correspondence, Visual Similarity, Multi-view reasoning, Functional Correspondence, Relative Depth, Object Localization and Counting.

\subsection{Full prompts of proprietary models.}
The prompt of GPT-4o, GPT-4o with CoT and GPT-5-Thinking are shown in Table \ref{tab:prompts_gpt} accordingly.

\section{More qualitative examples}
\label{sec:more_qualitative}
\subsection{More Visualization of Reasoning Traces.}
\begin{figure*}[t]
    \centering
    \includegraphics[width=\linewidth]{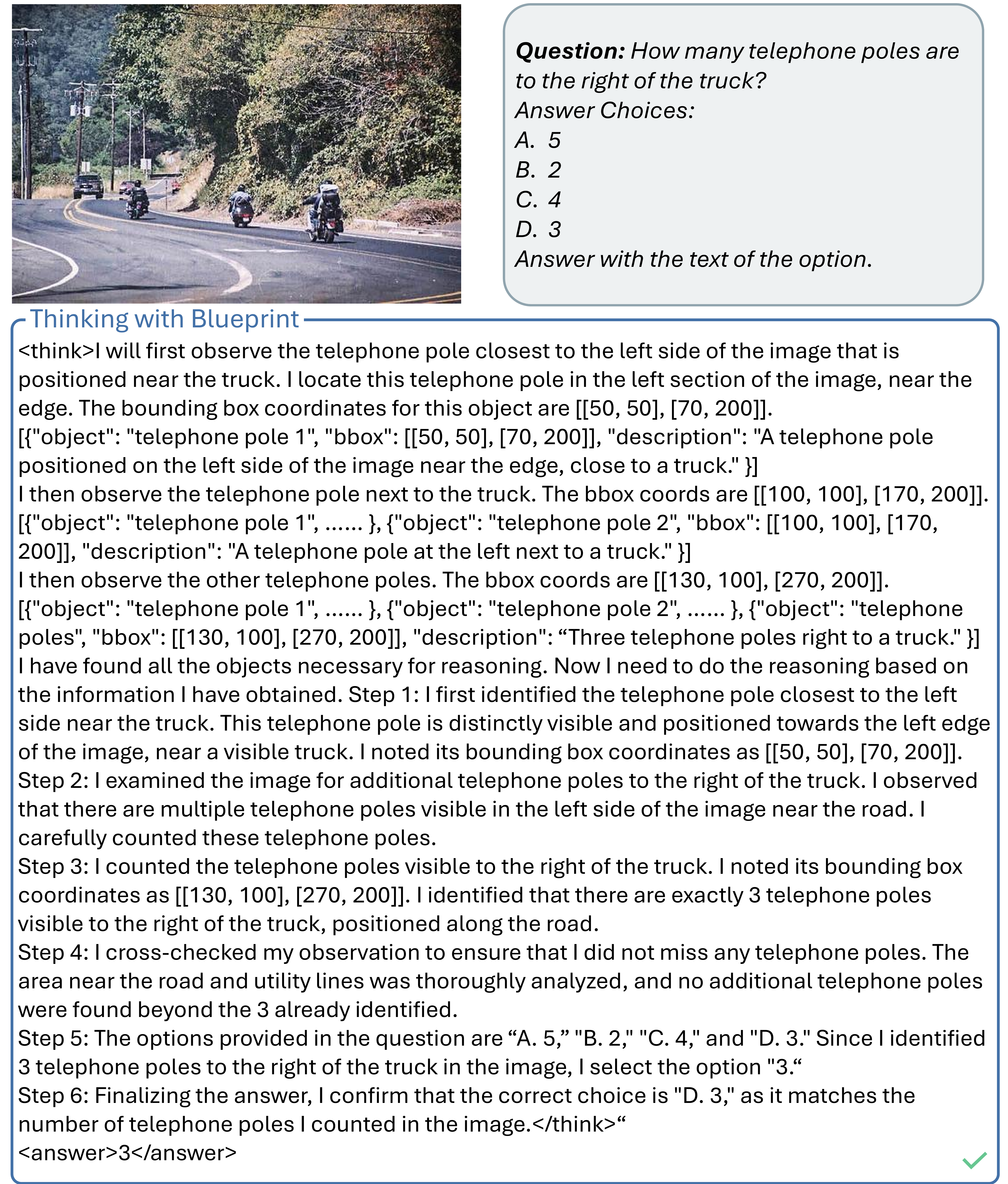}
    \caption{Qualitative example 1. (1 of 3).
    }
    \label{fig:qual_1}
\end{figure*}

\begin{figure*}[t]
    \centering
    \includegraphics[width=\linewidth]{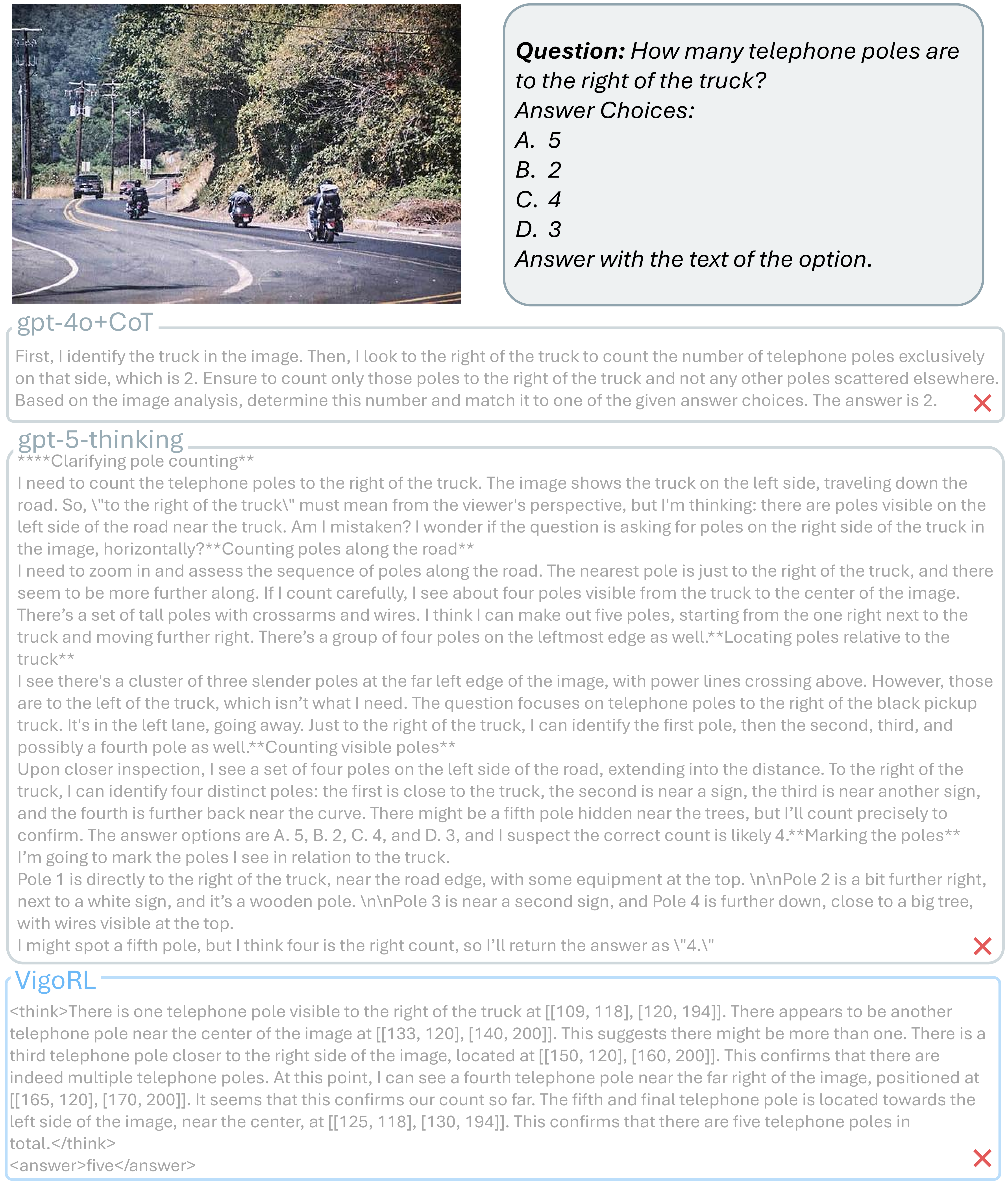}
    \caption{Qualitative example 1. (2 of 3).
    }
    \label{fig:qual_2}
\end{figure*}

\begin{figure*}[t]
    \centering
    \includegraphics[width=\linewidth]{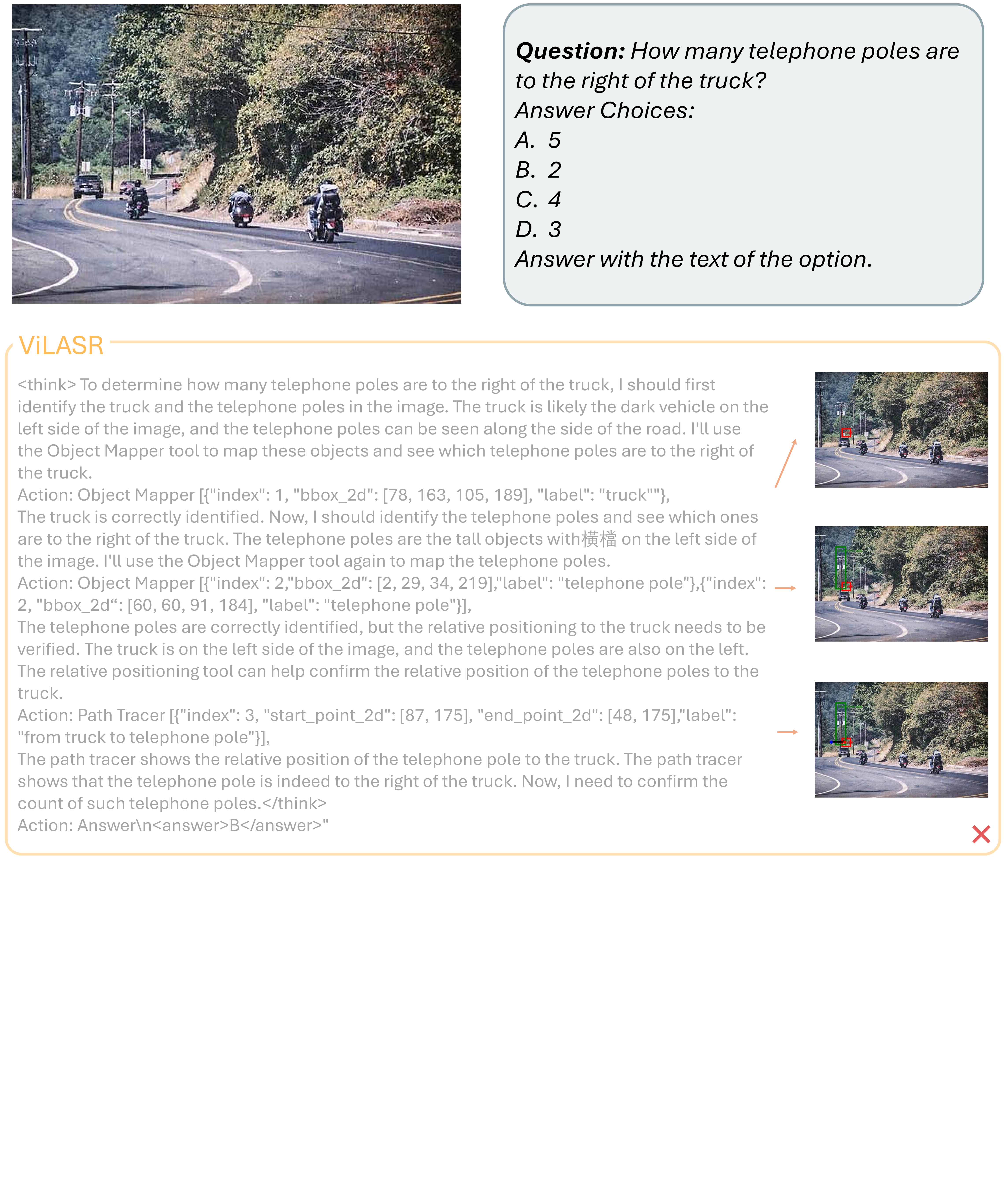}
    \caption{Qualitative example 1. (3 of 3).
    }
    \label{fig:qual_3}
\end{figure*}

\begin{figure*}[t]
    \centering
    \includegraphics[width=\linewidth]{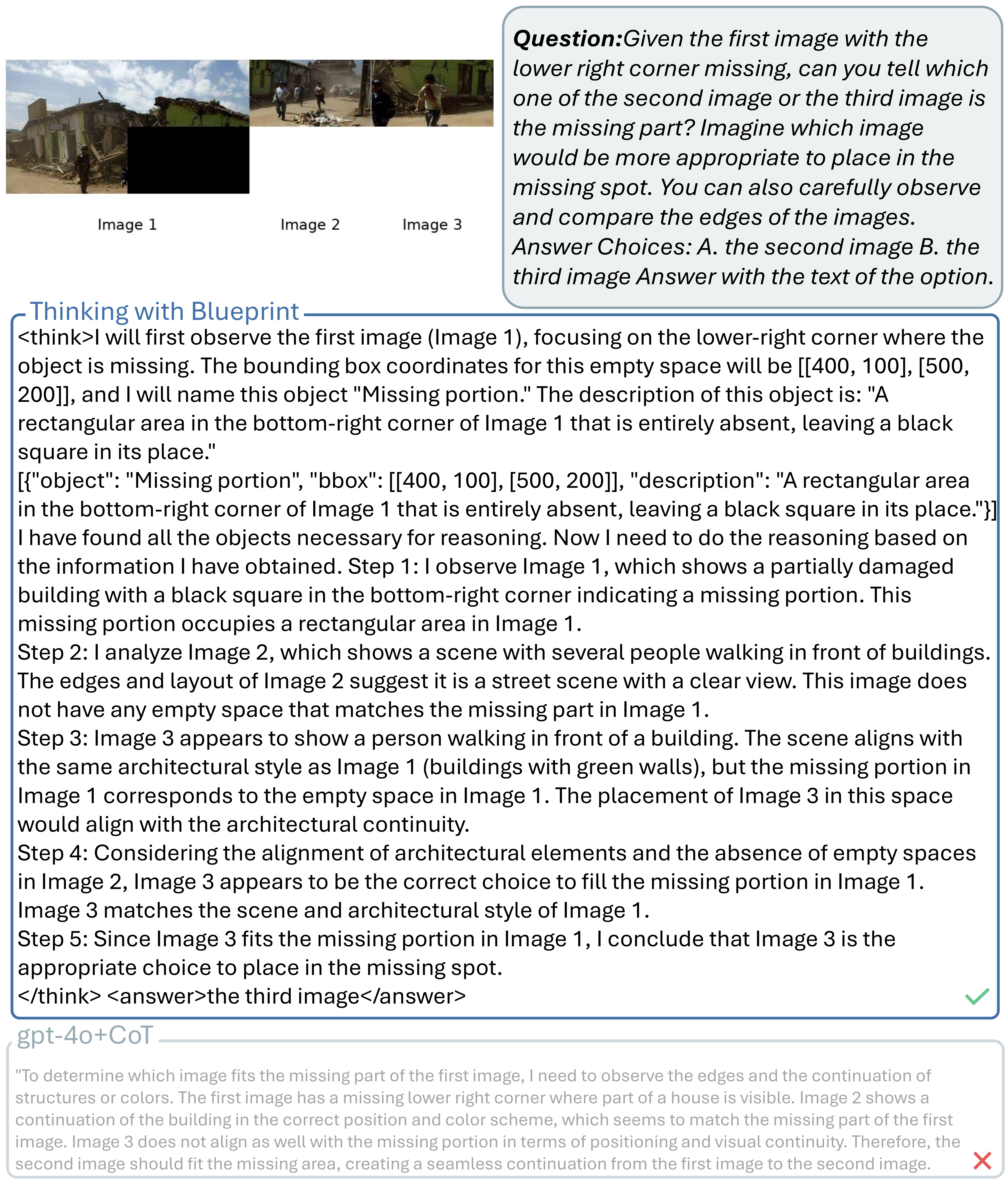}
    \caption{Qualitative example 2. (1 of 2).
    }
    \label{fig:qual_4}
\end{figure*}

\begin{figure*}[t]
    \centering
    \includegraphics[width=\linewidth]{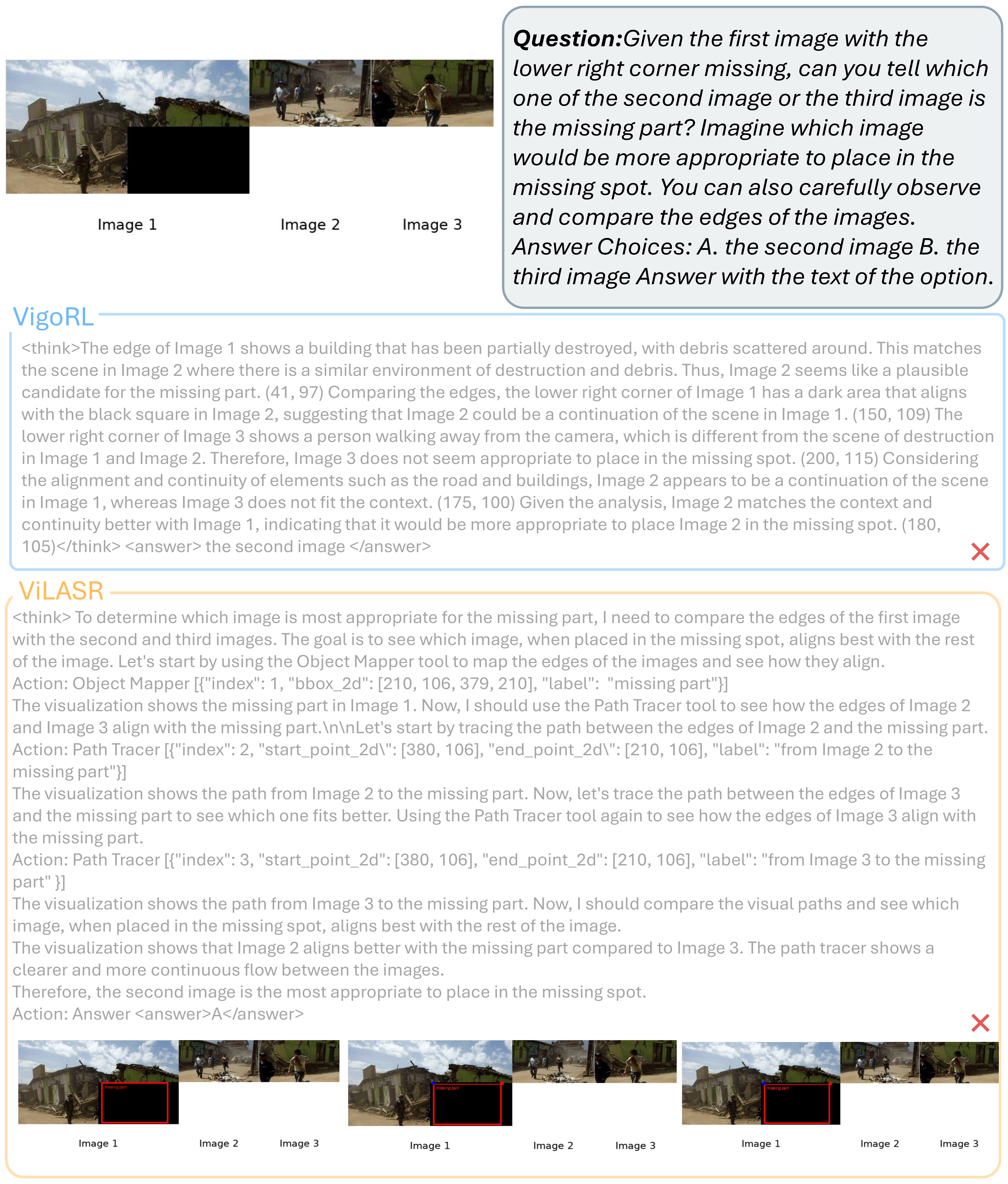}
    \caption{Qualitative example 2. (3 of 3).
    }
    \label{fig:qual_6}
\end{figure*}

\begin{figure*}[t]
    \centering
    \includegraphics[width=\linewidth]{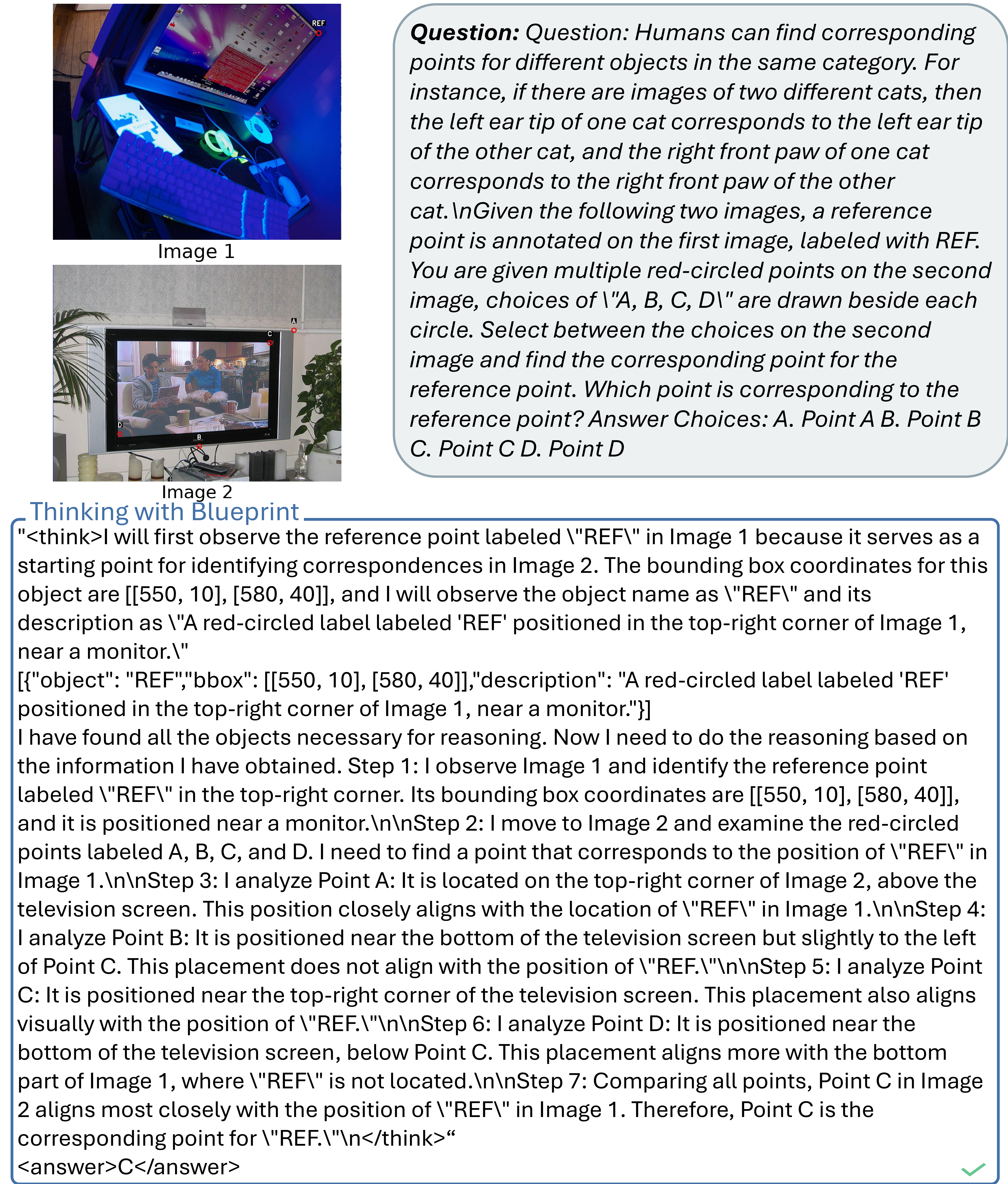}
    \caption{Qualitative example 3. (1 of 3).
    }
    \label{fig:qual_7}
\end{figure*}

\begin{figure*}[t]
    \centering
    \includegraphics[width=\linewidth]{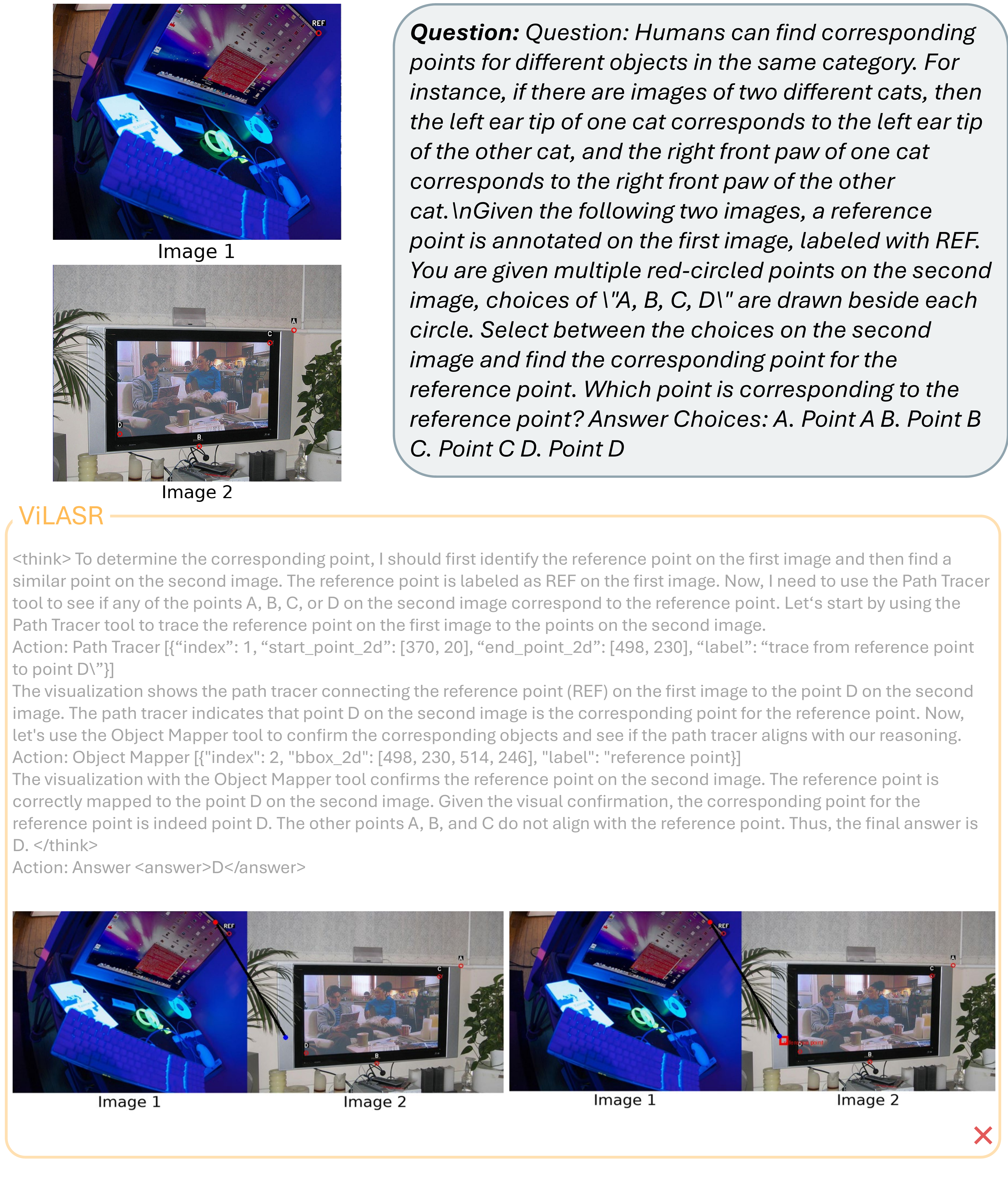}
    \caption{Qualitative example 3. (3 of 3).
    }
    \label{fig:qual_9}
\end{figure*}

\begin{figure*}[t]
    \centering
    \includegraphics[width=\linewidth]{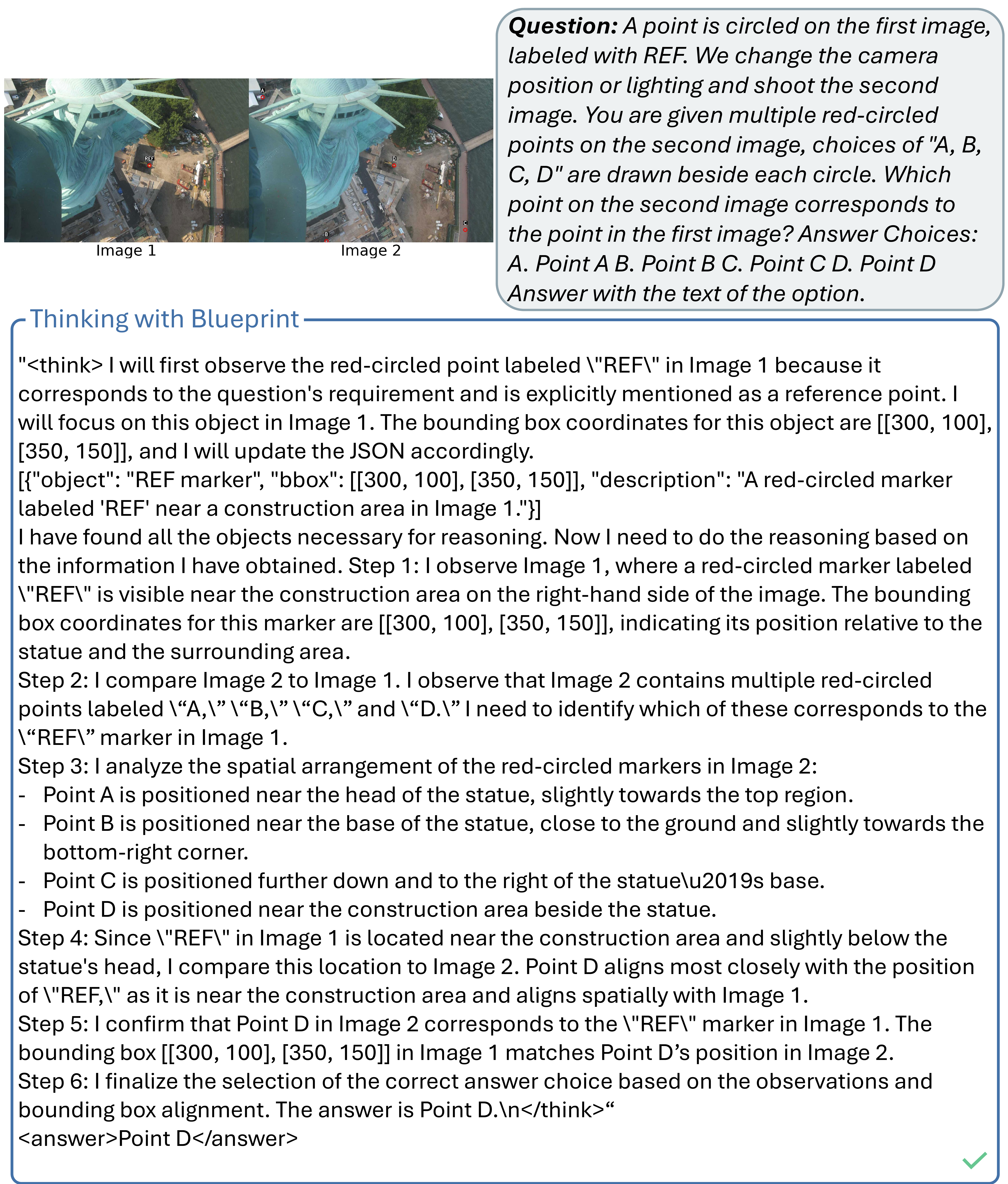}
    \caption{Qualitative example 4. (1 of 3).
    }
    \label{fig:qual_10}
\end{figure*}

\begin{figure*}[t]
    \centering
    \includegraphics[width=\linewidth]{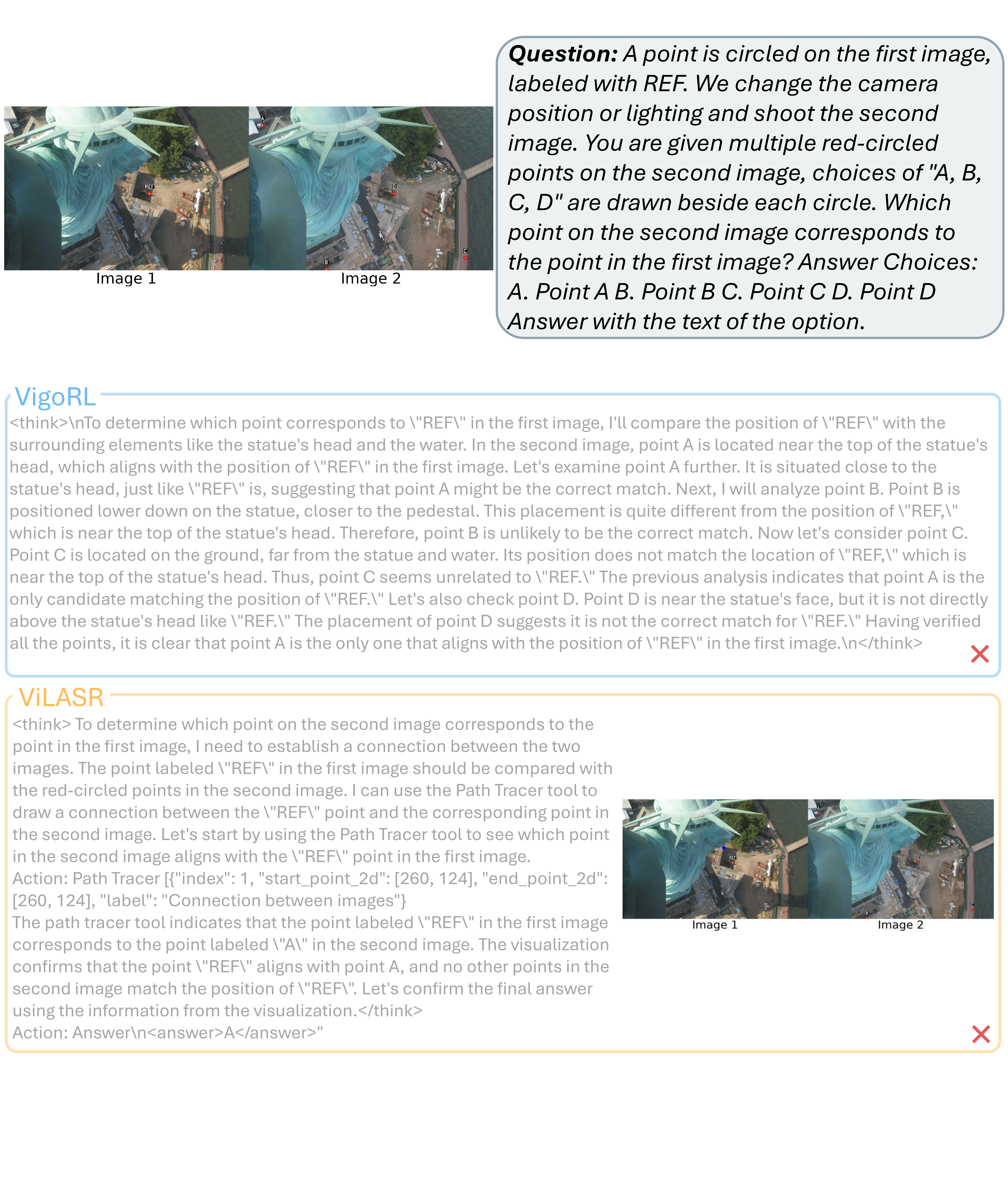}
    \caption{Qualitative example 4. (3 of 3).
    }
    \label{fig:qual_12}
\end{figure*}

\begin{figure*}[t]
    \centering
    \includegraphics[width=\linewidth]{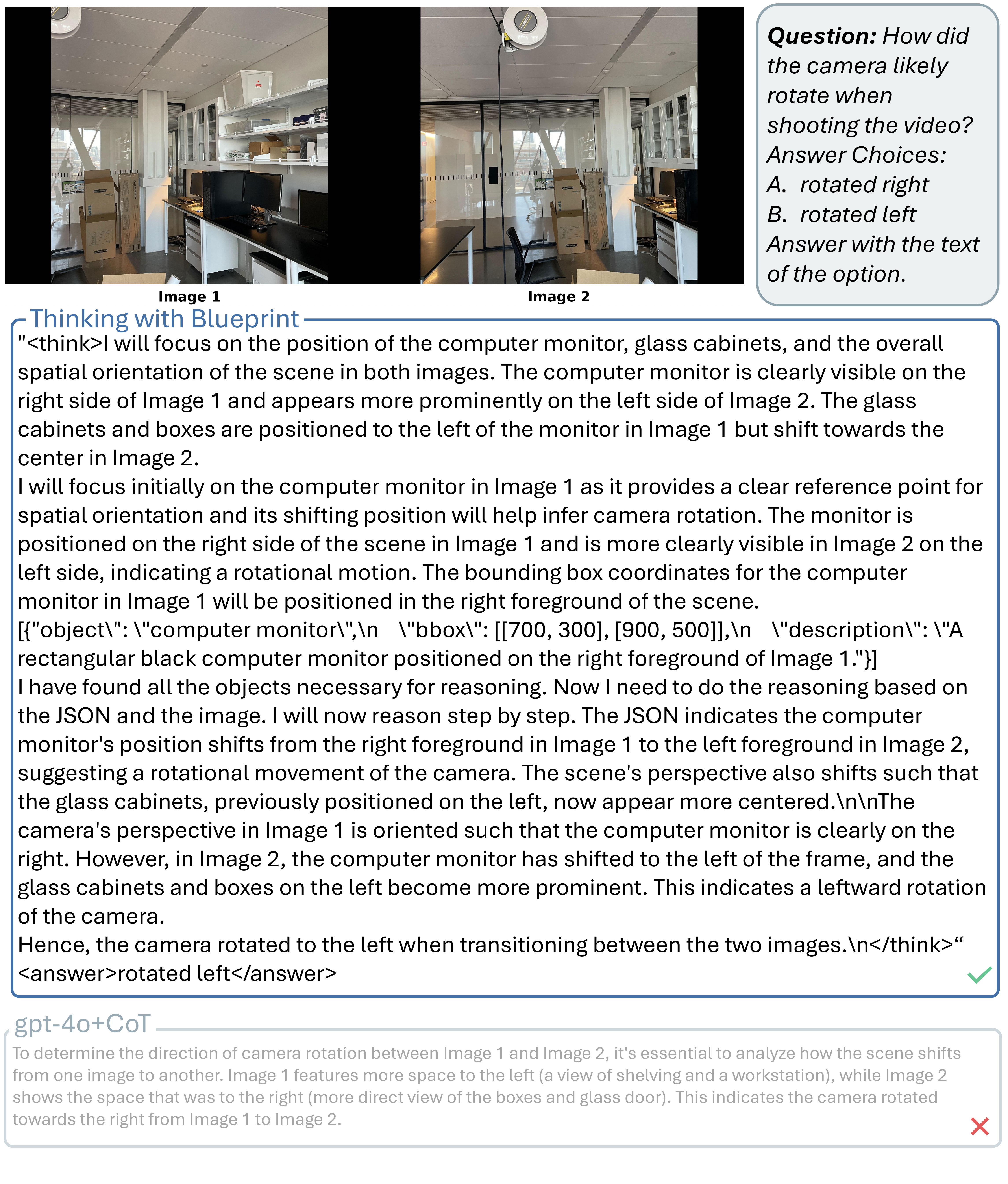}
    \caption{Qualitative example 5. (1 of 3).
    }
    \label{fig:qual_13}
\end{figure*}

\begin{figure*}[t]
    \centering
    \includegraphics[width=\linewidth]{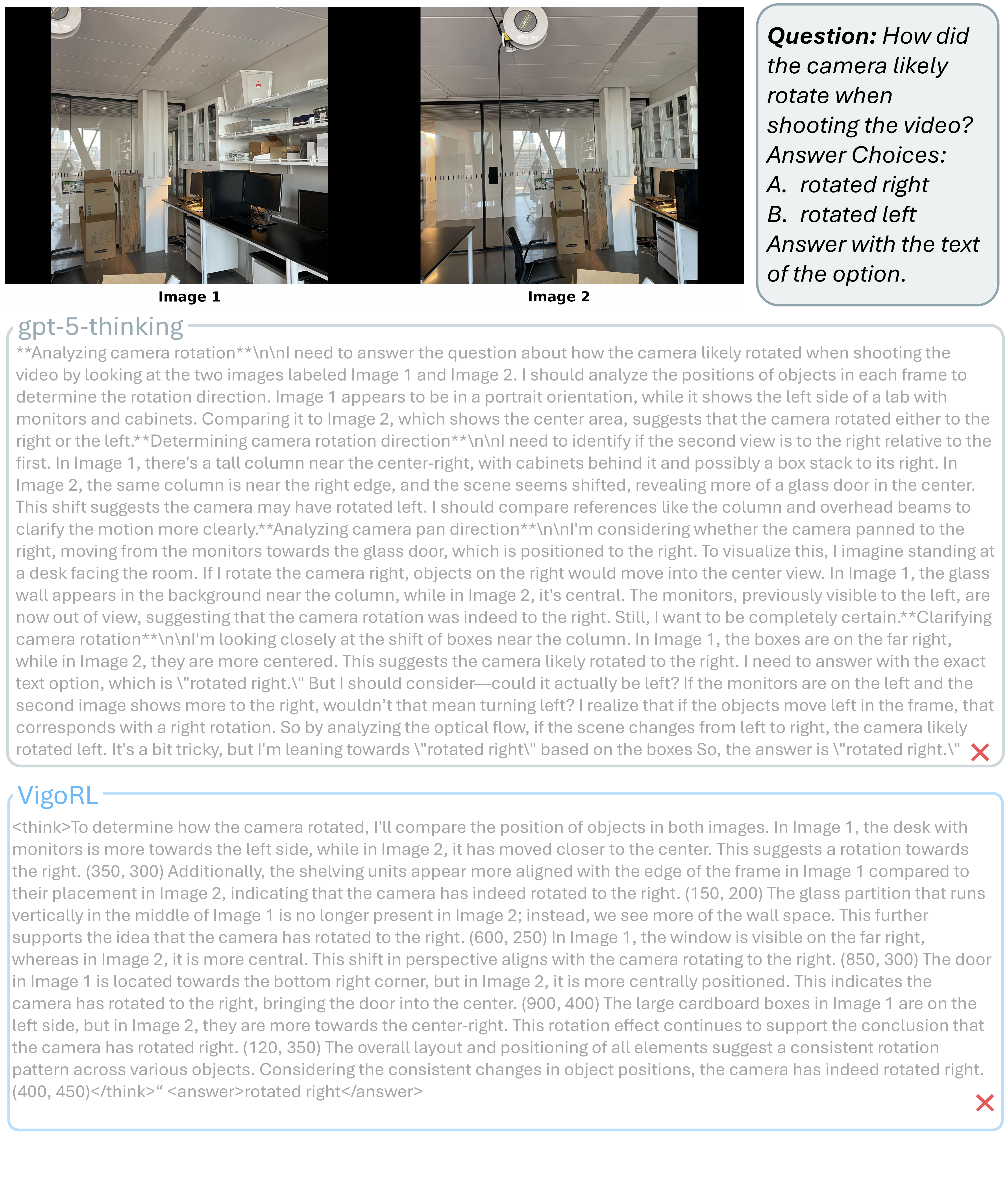}
    \caption{Qualitative example 5. (2 of 3).
    }
    \label{fig:qual_14}
\end{figure*}

\begin{figure*}[t]
    \centering
    \includegraphics[width=\linewidth]{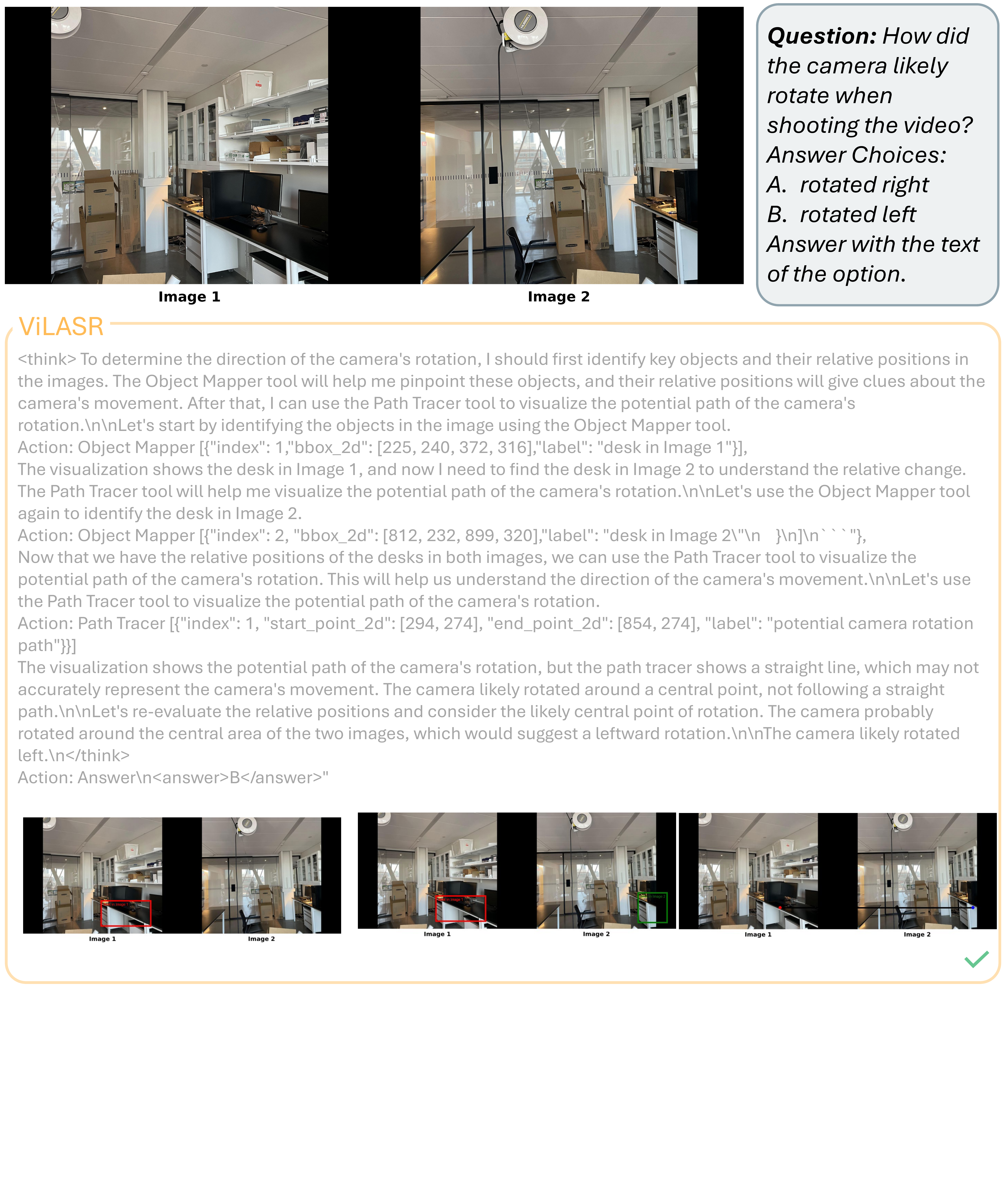}
    \caption{Qualitative example 5. (3 of 3).
    }
    \label{fig:qual_15}
\end{figure*}

\begin{figure*}[t]
    \centering
    \includegraphics[width=\linewidth]{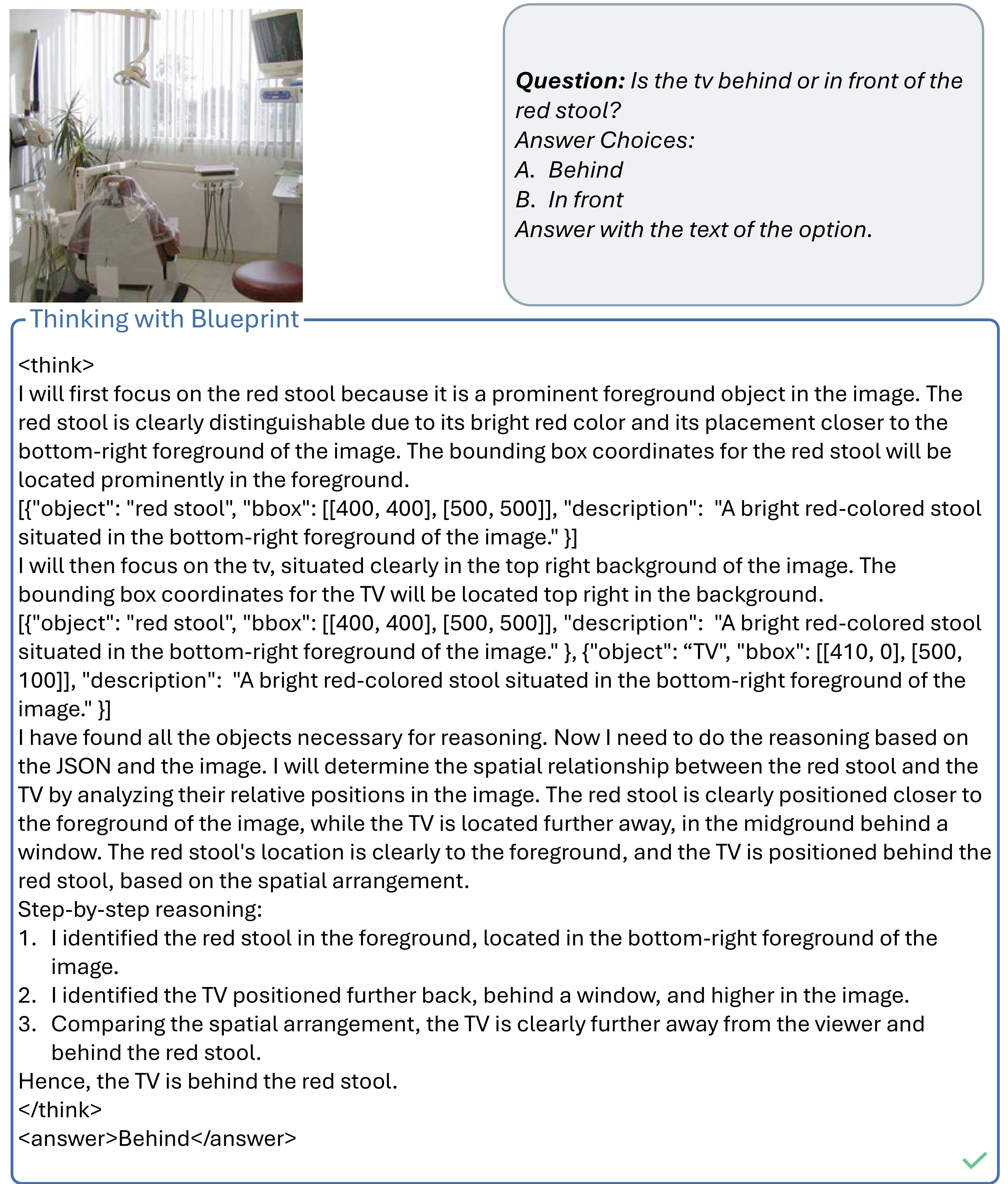}
    \caption{Qualitative example 6. (1 of 2).
    }
    \label{fig:qual_16}
\end{figure*}

\begin{figure*}[t]
    \centering
    \includegraphics[width=\linewidth]{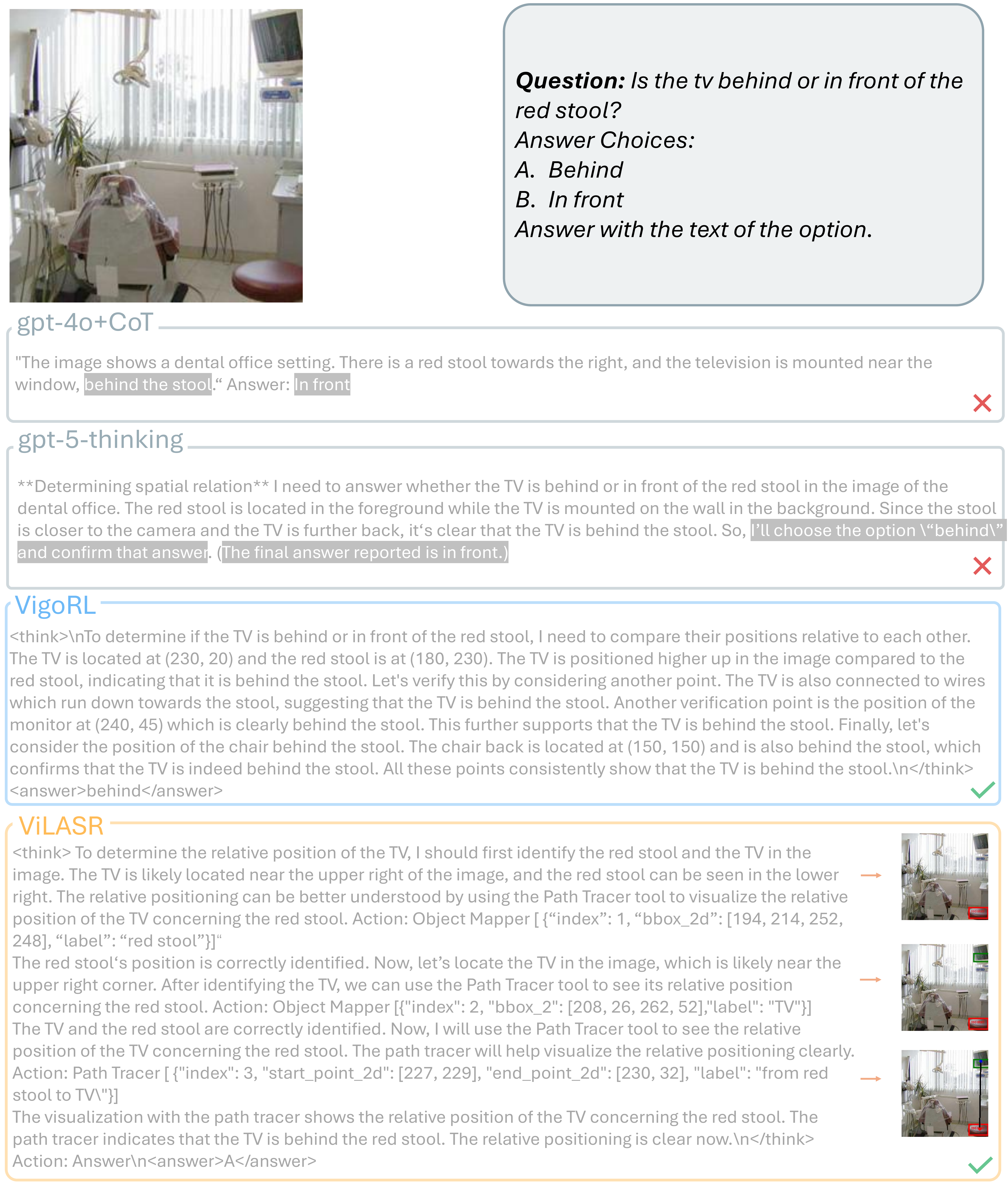}
    \caption{Qualitative example 6. (2 of 2). We can observe inconsistency between reasoning and answer in both gpt-4o + CoT and gpt-5-thinking.
    }
    \label{fig:qual_17}
\end{figure*}

\begin{figure*}[t]
    \centering
    \includegraphics[width=\linewidth]{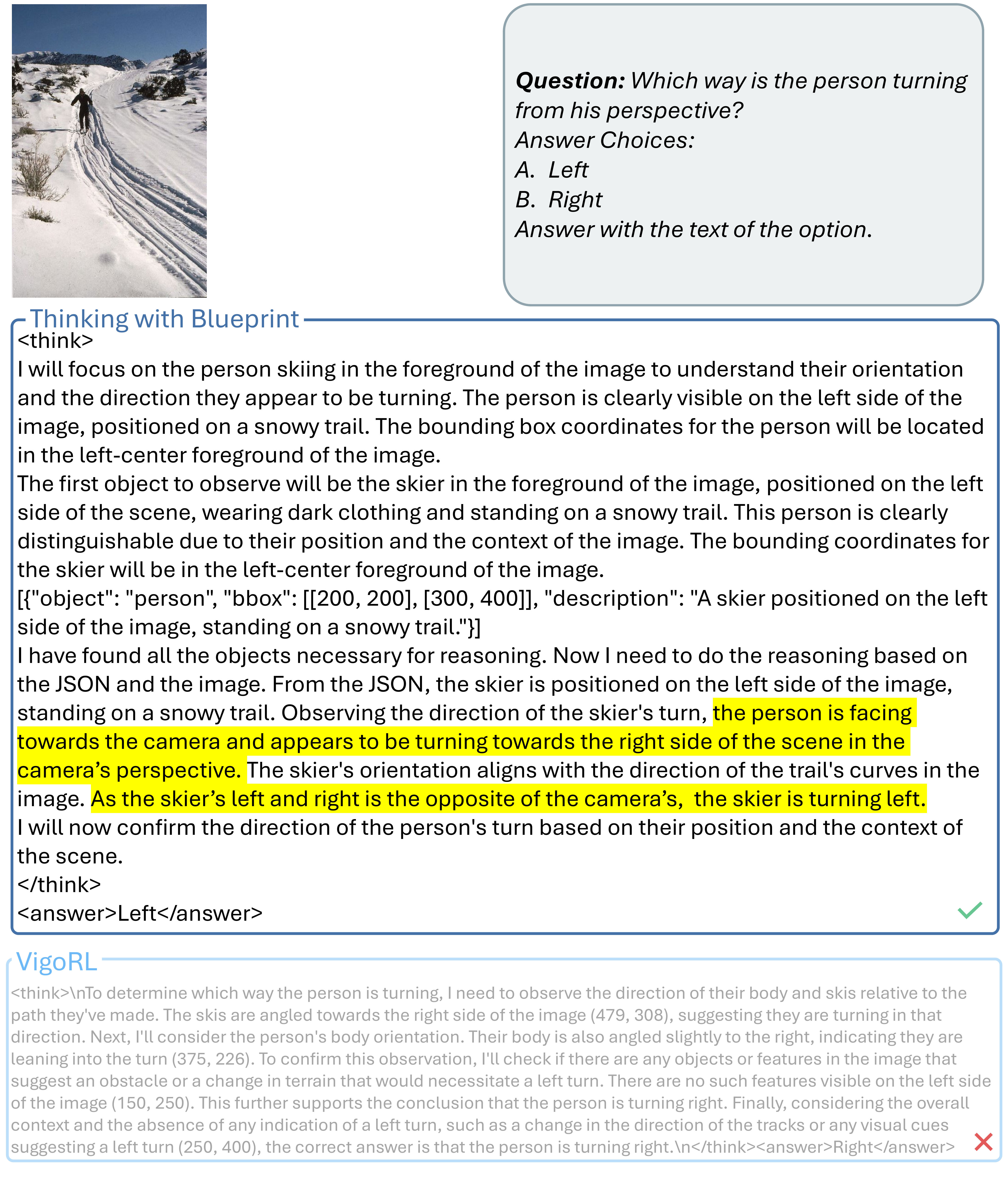}
    \caption{Qualitative example 7. (1 of 2).
    }
    \label{fig:qual_18}
\end{figure*}

\begin{figure*}[t]
    \centering
    \includegraphics[width=\linewidth]{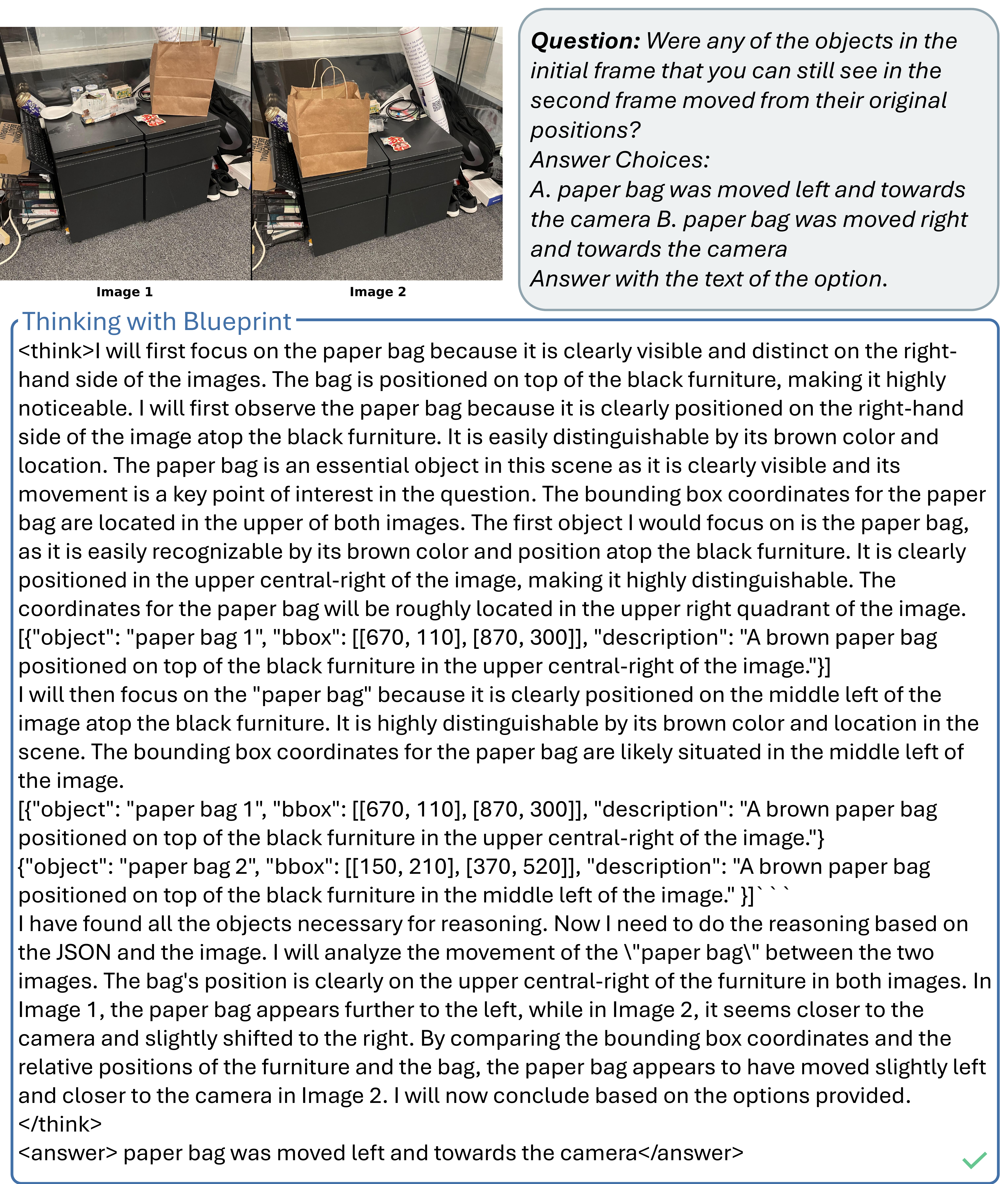}
    \caption{Qualitative example 8. (1 of 3).
    }
    \label{fig:qual_20}
\end{figure*}

\begin{figure*}[t]
    \centering
    \includegraphics[width=\linewidth]{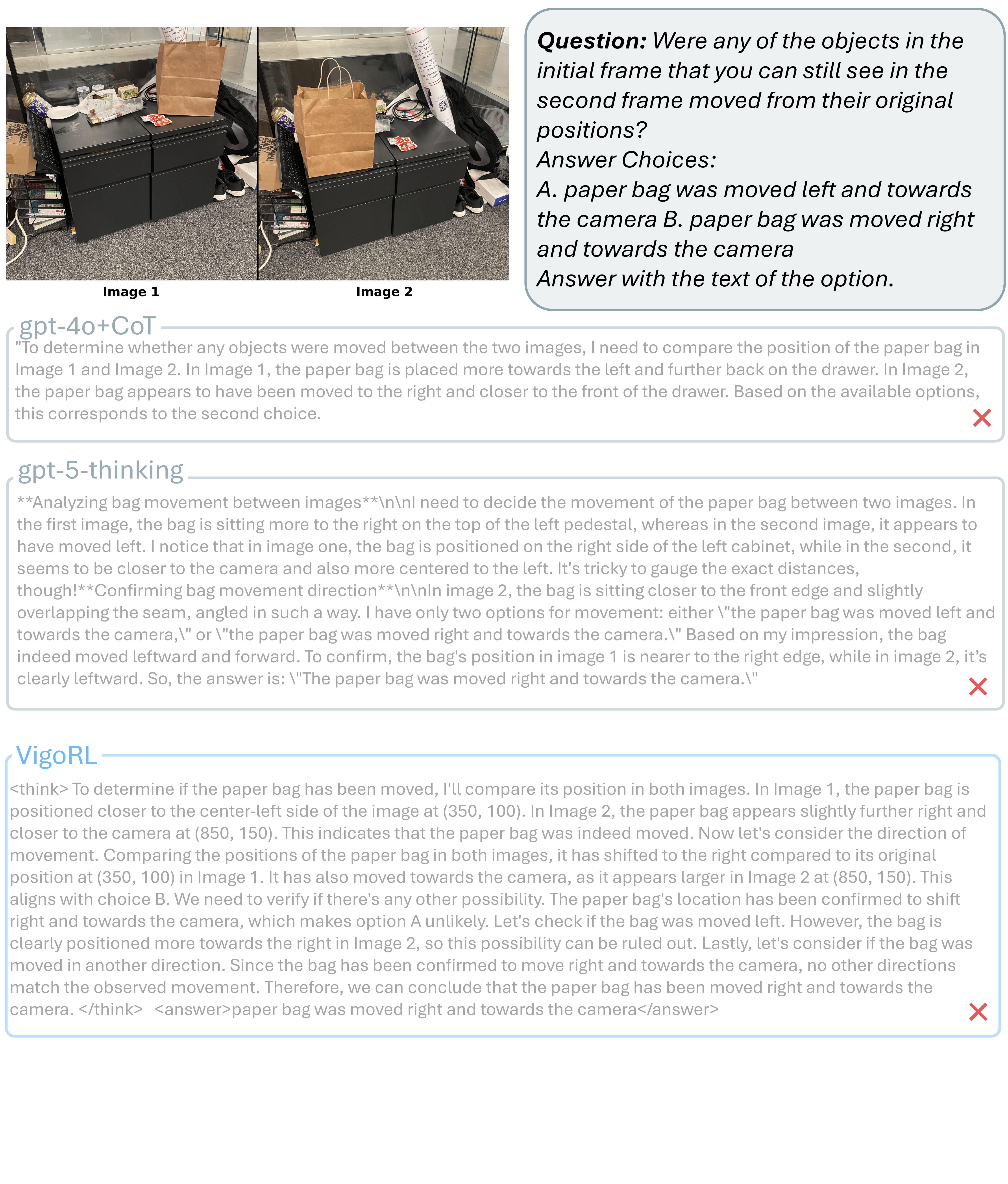}
    \caption{Qualitative example 8. (2 of 3).
    }
    \label{fig:qual_21}
\end{figure*}

\begin{figure*}[t]
    \centering
    \includegraphics[width=\linewidth]{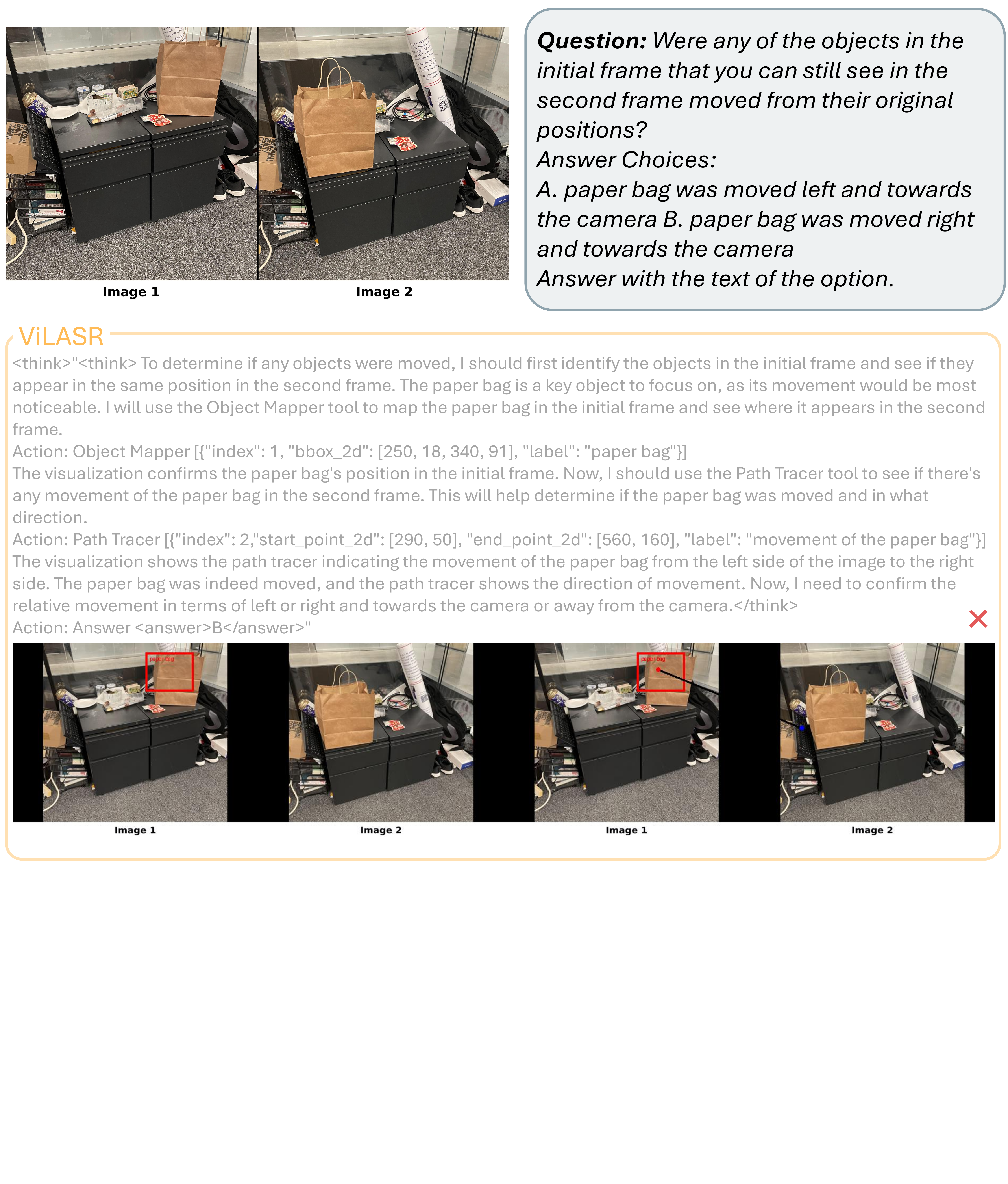}
    \caption{Qualitative example 8. (3 of 3).
    }
    \label{fig:qual_22}
\end{figure*}

\begin{figure*}[t]
    \centering
    \includegraphics[width=\linewidth]{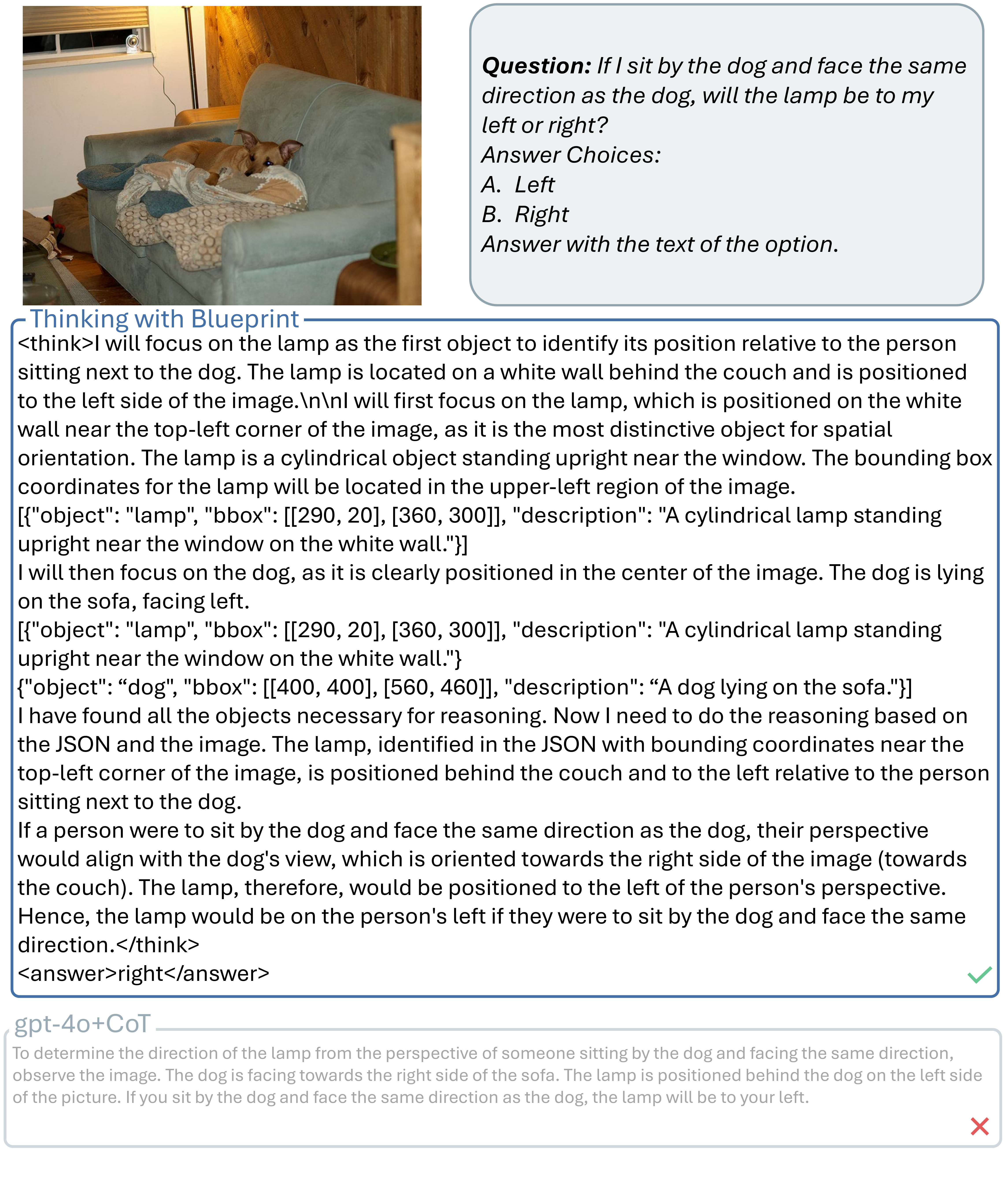}
    \caption{Qualitative example 9. (1 of 3).
    }
    \label{fig:qual_23}
\end{figure*}

\begin{figure*}[t]
    \centering
    \includegraphics[width=\linewidth]{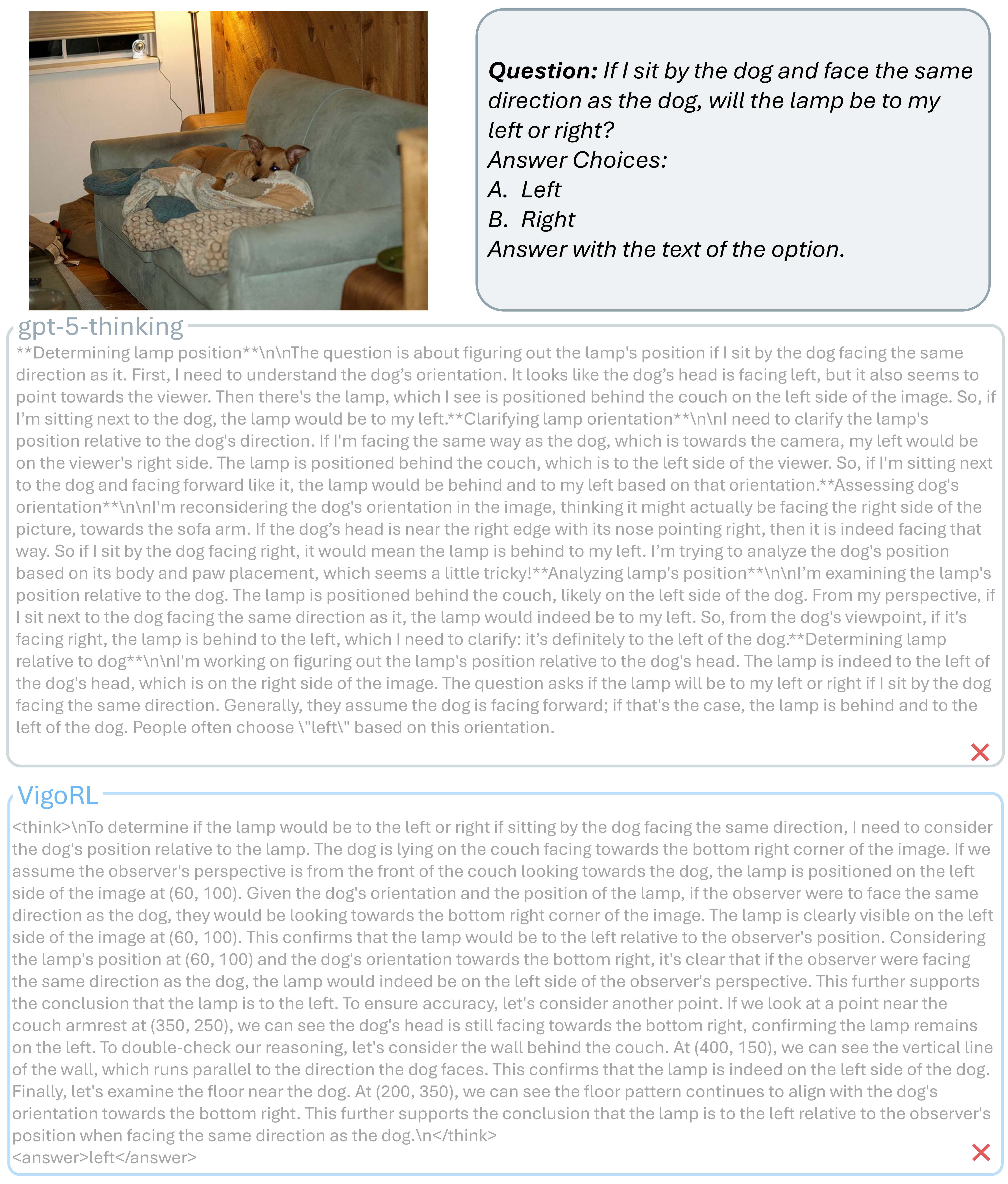}
    \caption{Qualitative example 9. (2 of 3).
    }
    \label{fig:qual_24}
\end{figure*}

\begin{figure*}[t]
    \centering
    \includegraphics[width=\linewidth]{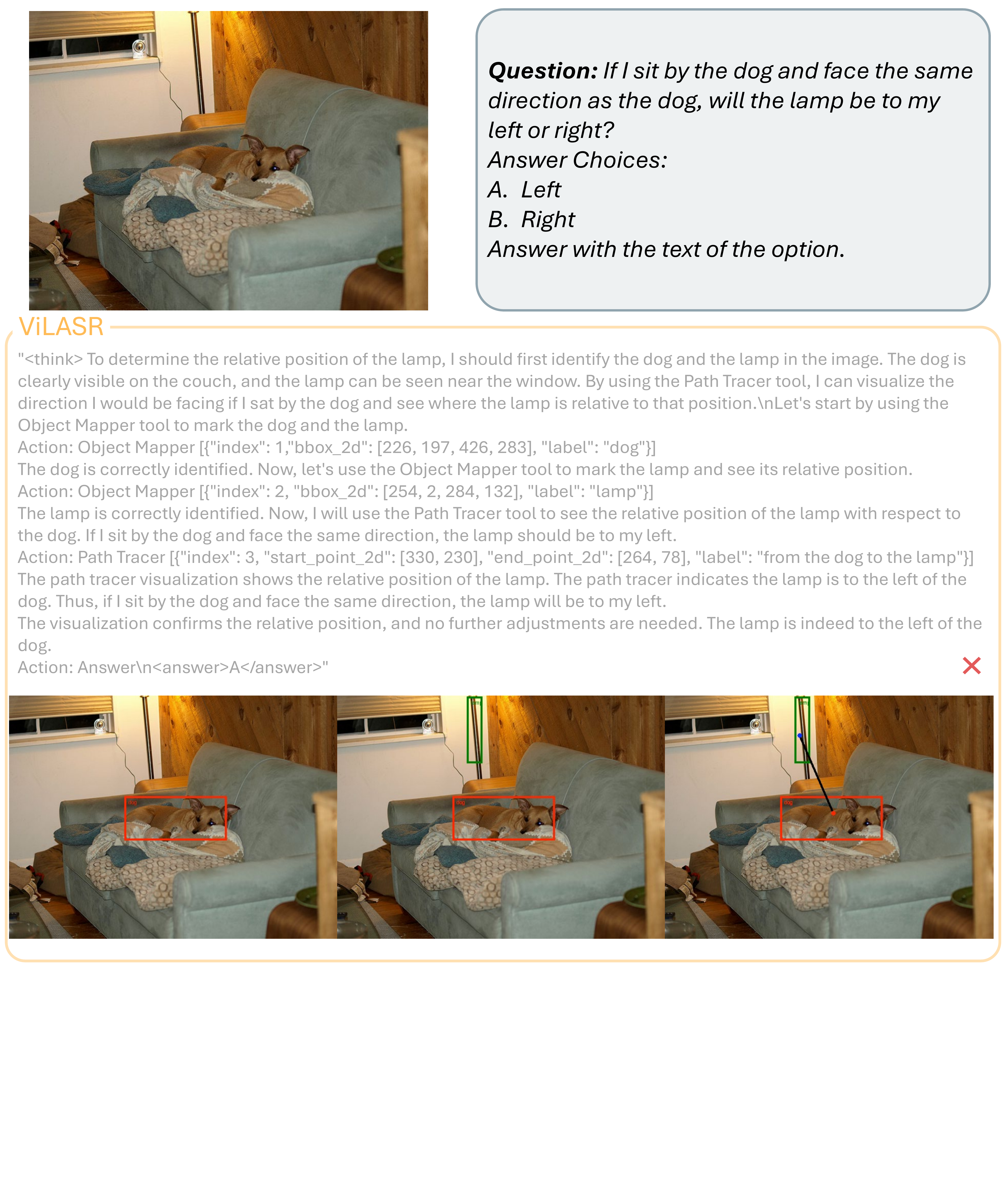}
    \caption{Qualitative example 9. (3 of 3).
    }
    \label{fig:qual_25}
\end{figure*}

\begin{figure*}[t]
    \centering
    \includegraphics[width=\linewidth]{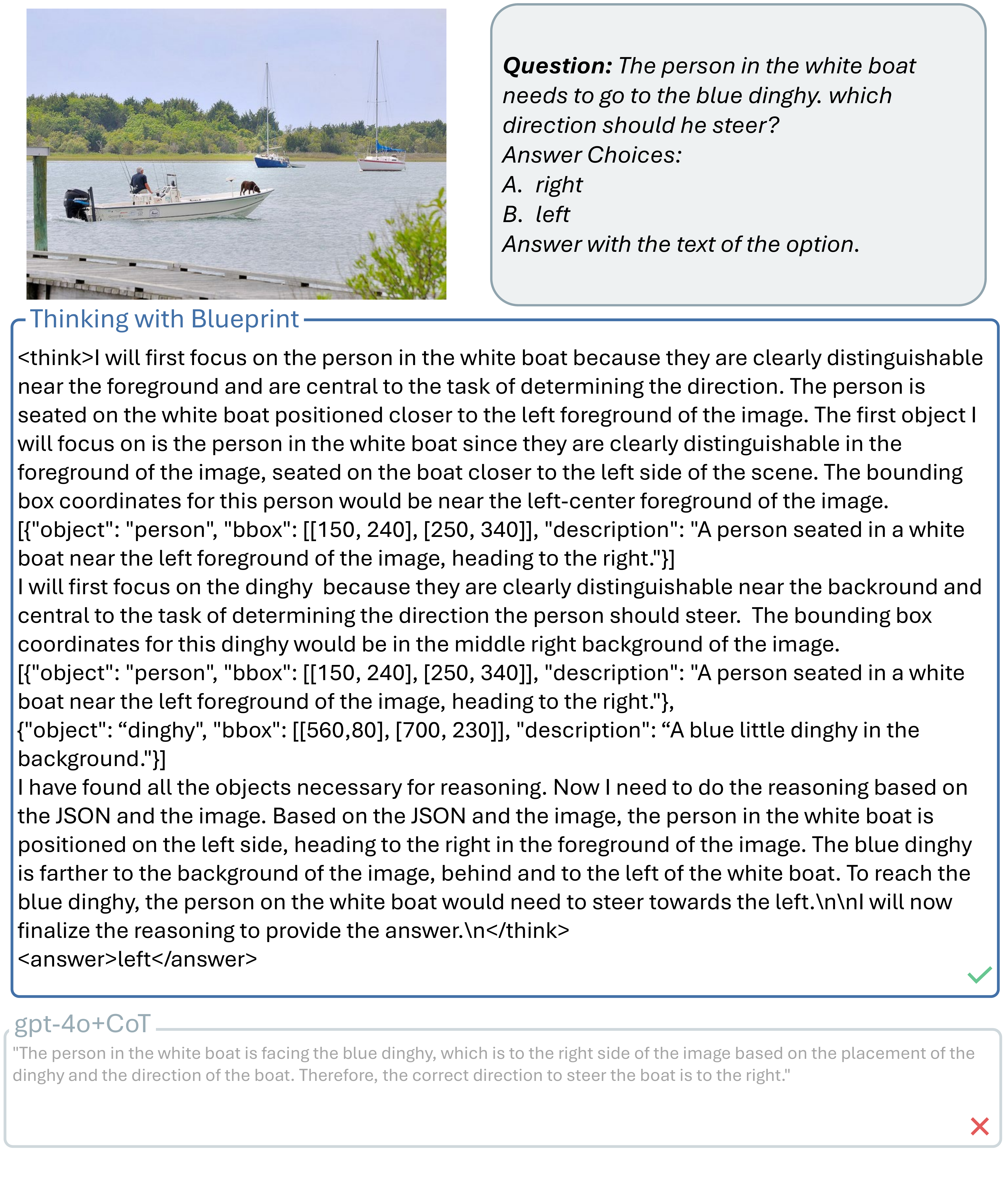}
    \caption{Qualitative example 10. (1 of 3).
    }
    \label{fig:qual_26}
\end{figure*}

\begin{figure*}[t]
    \centering
    \includegraphics[width=\linewidth]{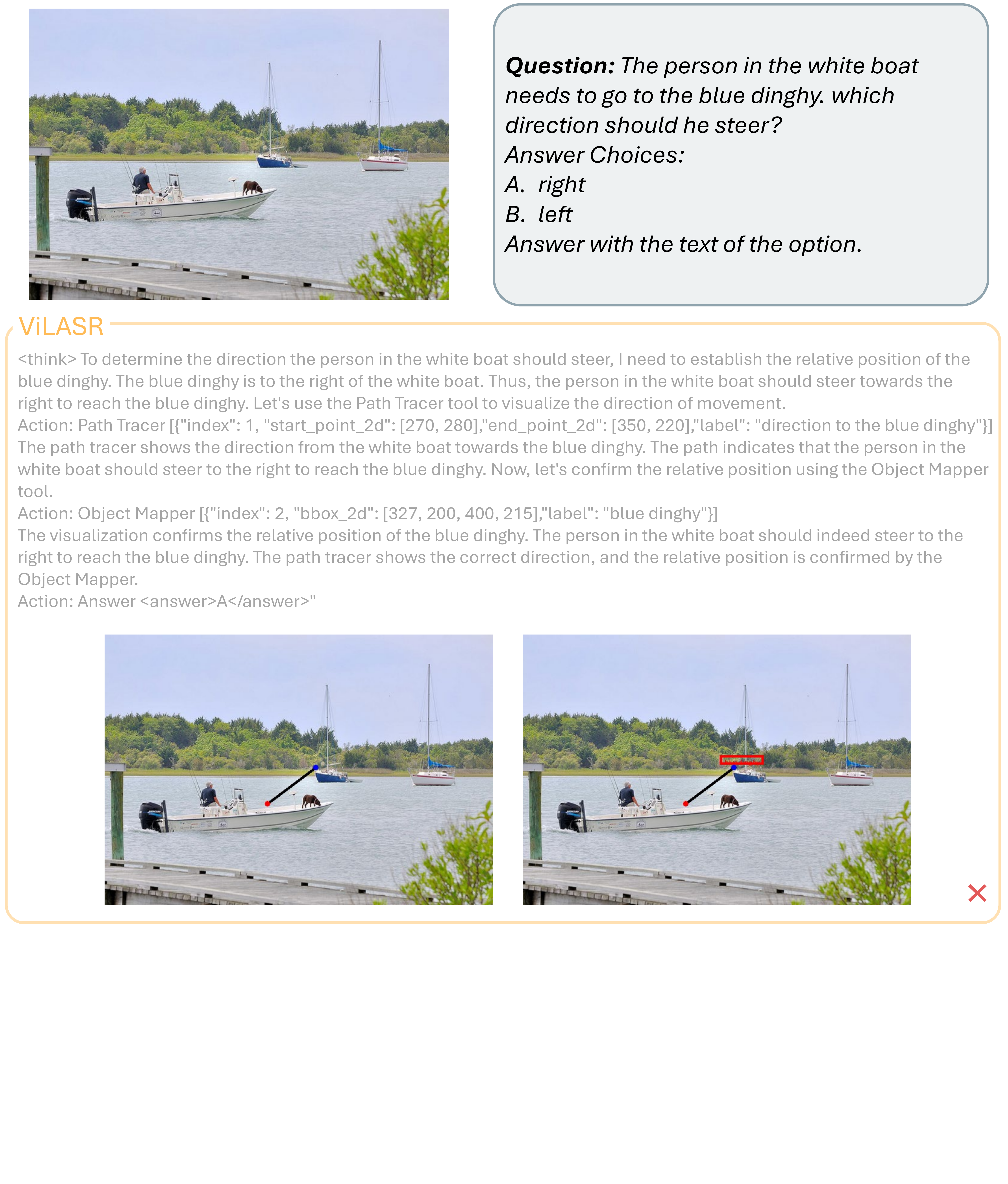}
    \caption{Qualitative example 10. (3 of 3).
    }
    \label{fig:qual_28}
\end{figure*}

More visualization of the whole reasoning trajectory of different methods is shown from Figure \ref{fig:qual_1} to Figure \ref{fig:qual_28}. From the qualitative examples we can discover that our method yields far better performance in different kinds of spatial reasoning tasks, thanks to the blueprint-based thinking patterns introduced, as well as the corresponding training strategies designed.

It is worth noting that we also discovered the inconsistency between reasoning and final answer in gpt-4o and gpt-5-thinking (Figure \ref{fig:qual_17}), illustrating that the inconsistency issue is agnostic among base models, thus further illustrating the importance of our design of the consistency reward.

\subsection{About Video Spatial Reasoning.}
\noindent\textbf{Method details for adapting our method to video spatial reasoning.}
Here we detail about how to combine our method with a frame selector so as to expand our method to video-based spatial reasoning. Following Q-Frame, we first sample 128 frames uniformly from the input video. SigLIP is then applied to re-rank the downsampled 128 frames to select the top 4 candidate frames. Our model goes through the frames one by one. If the frame does not have the relevant information, it will say \textit{"Questions and image do not match."} and go to the next frame. Our model stops when it finds a frame where all the information about the question is inside the frame, or when all four candidate frames are used up but it did not find the proper answer. 

\begin{figure*}[t]
    \centering
    \includegraphics[width=\linewidth]{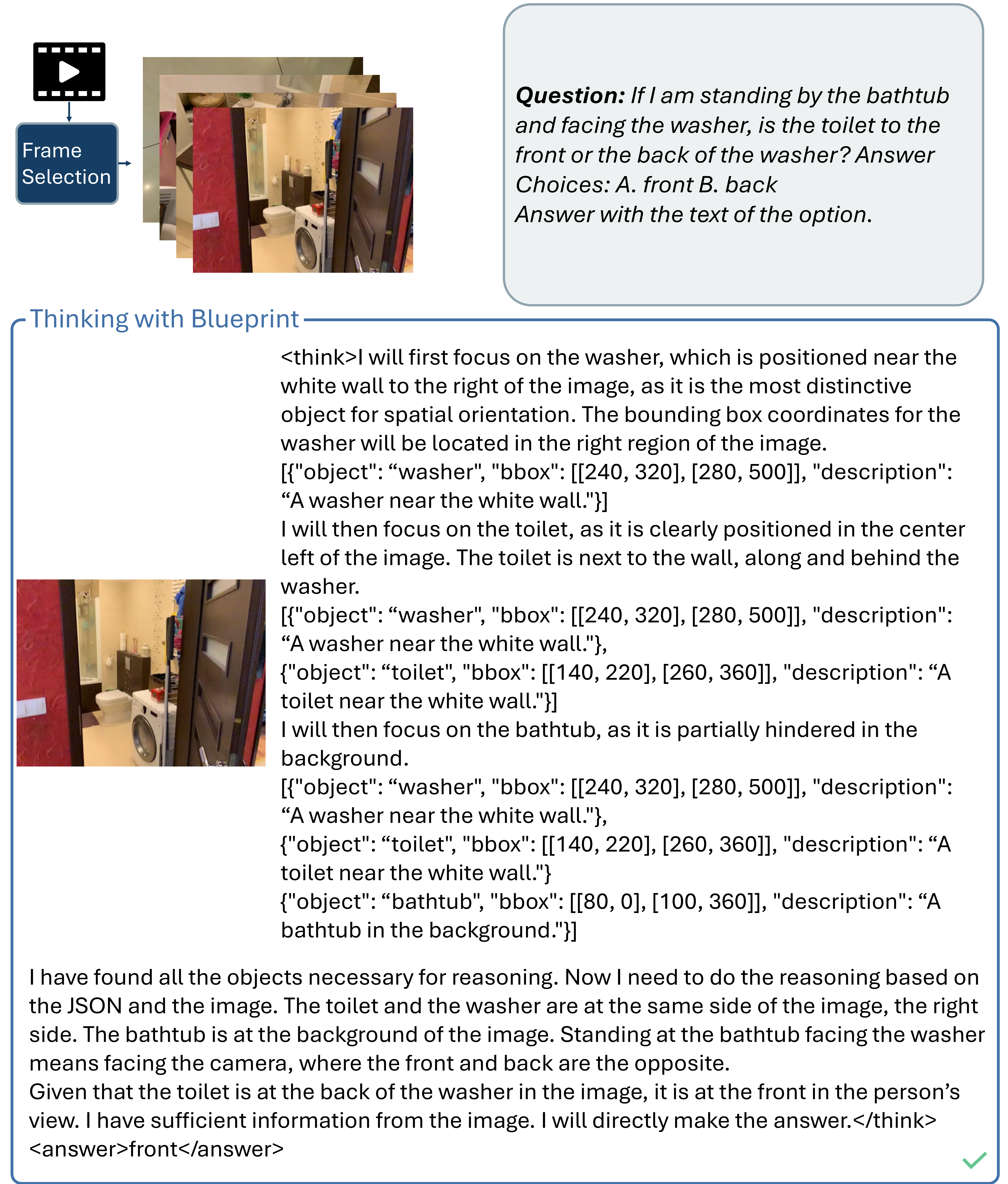}
    \caption{Qualitative example of appyling our method to video spatial reasoning.
    }
    \label{fig:vsi_1}
\end{figure*}

\begin{figure*}[t]
    \centering
    \includegraphics[width=\linewidth]{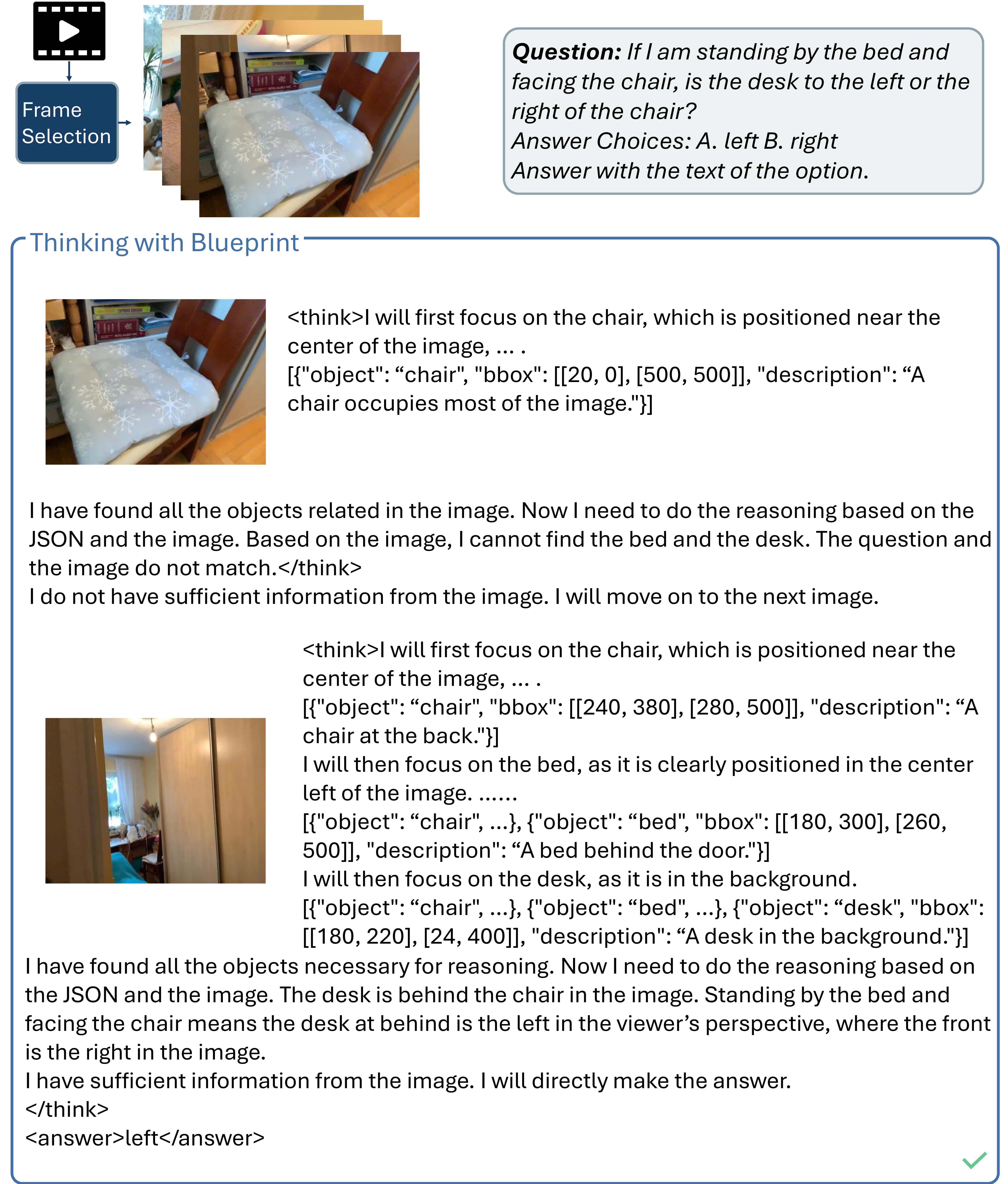}
    \caption{Qualitative example of appyling our method to video spatial reasoning.
    }
    \label{fig:vsi_2}
\end{figure*}

\noindent\textbf{More Qualitative Results of Video Spatial Reasoning.}
The qualitative illustrations about how to extend our method to video spatial reasoning is shown in Figure \ref{fig:vsi_1} and Figure \ref{fig:vsi_2}. From Figure \ref{fig:vsi_1} we can discover that our model can quickly reach the correct answer when the correct frame is on top of the candidate list. Figure \ref{fig:vsi_2} also showcases that our model can exclude irrelevant frames when Q-Frames fail to rank the most relevant one on top of the list and finally navigate itself to the correct frame. Thanks to the methodological design in the adversarial images, which extends the model's capability to distinguish irrelevant frames and push our model further to video spatial understanding.

\subsection{Failure cases.}
In this part we showcase some typical failure cases. Failure cases mostly happen at the inability to perceive all the objects in the image, lacking the world knowledge prior, as well as the ignorance of the consecitive 3D space and the camera geometry. 

\begin{figure*}[t]
    \centering
    \includegraphics[width=\linewidth]{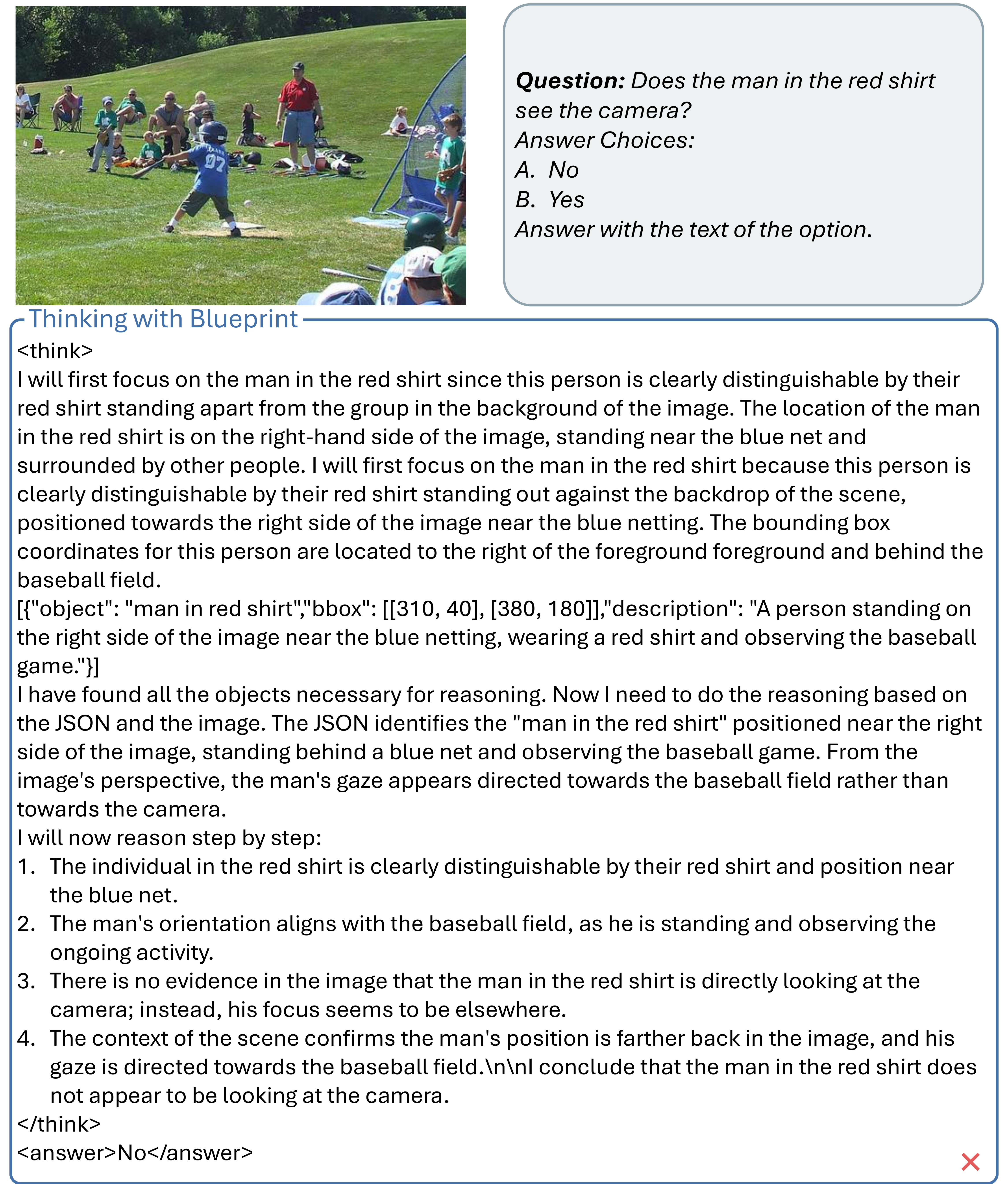}
    \caption{Failure example 1. Our model is distracted by the standing man and fails to see the man in red, wearing sunglasses, sitting in the background
    }
    \label{fig:fail_1}
\end{figure*}

\begin{figure*}[t]
    \centering
    \includegraphics[width=\linewidth]{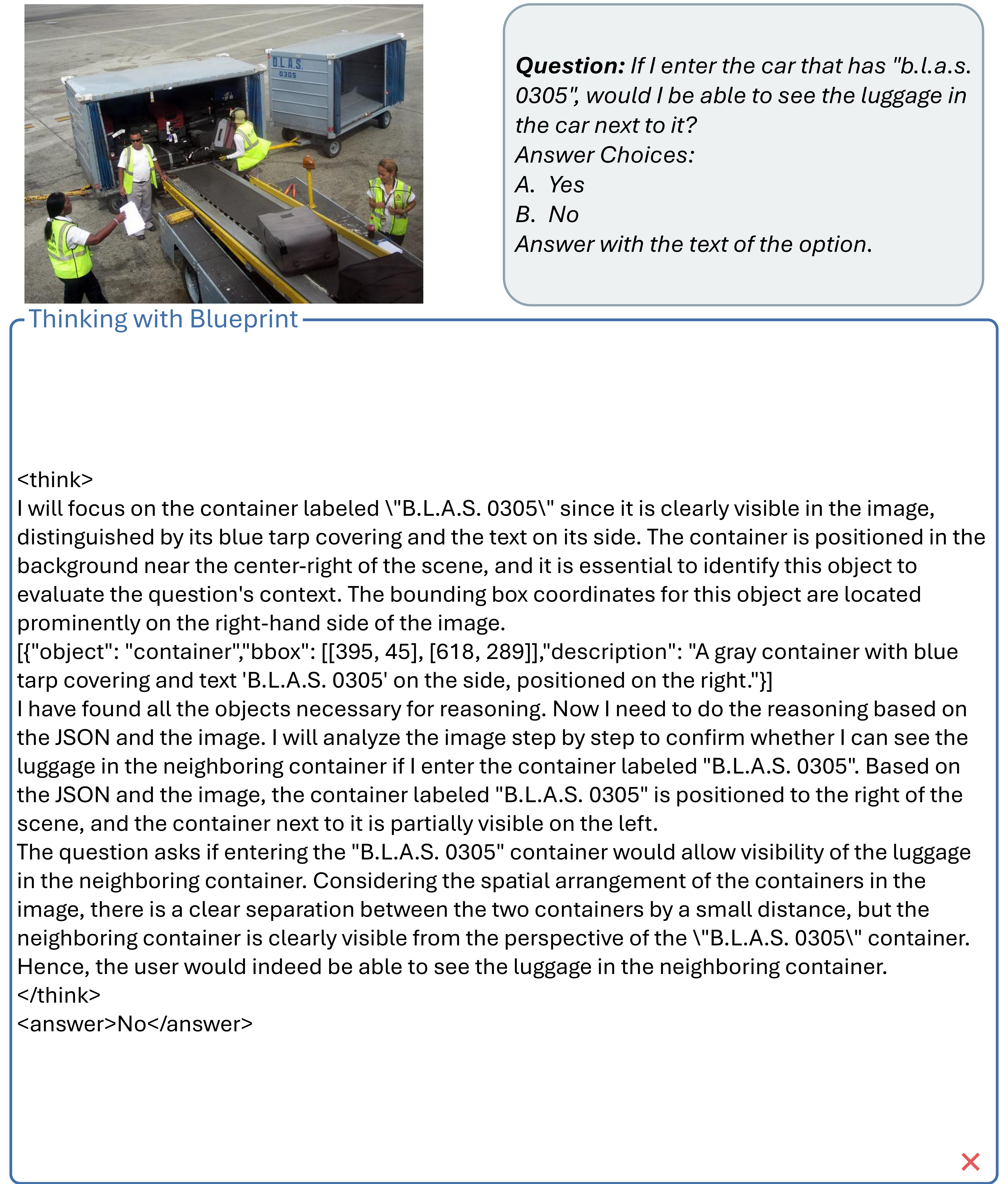}
    \caption{Failure example 2. Our model fails to discover that the first wagon is not transparent and cannot see the luggage in the second wagon from inside the metal cabin.
    }
    \label{fig:fail_2}
\end{figure*}

From Figure \ref{fig:fail_1} we can observe that failures may happen when our method fails to observe all the objects in the input image. In this example, for the man in red, our model only sees the standing man which is obvious in the foreground, but it has neglected the man sitting far away in the background, looking towards the camera behind his sunglasses. This calls for the stronger grounding capabilities in VLMs. Figure \ref{fig:fail_2} shows the case of the absence of real-world priors, where the model fails to realize that the sight of the luggages will be hindered by the non-transparent shells of the wagons if it entered the first wagon. 

\begin{figure*}[t]
    \centering
    \includegraphics[width=\linewidth]{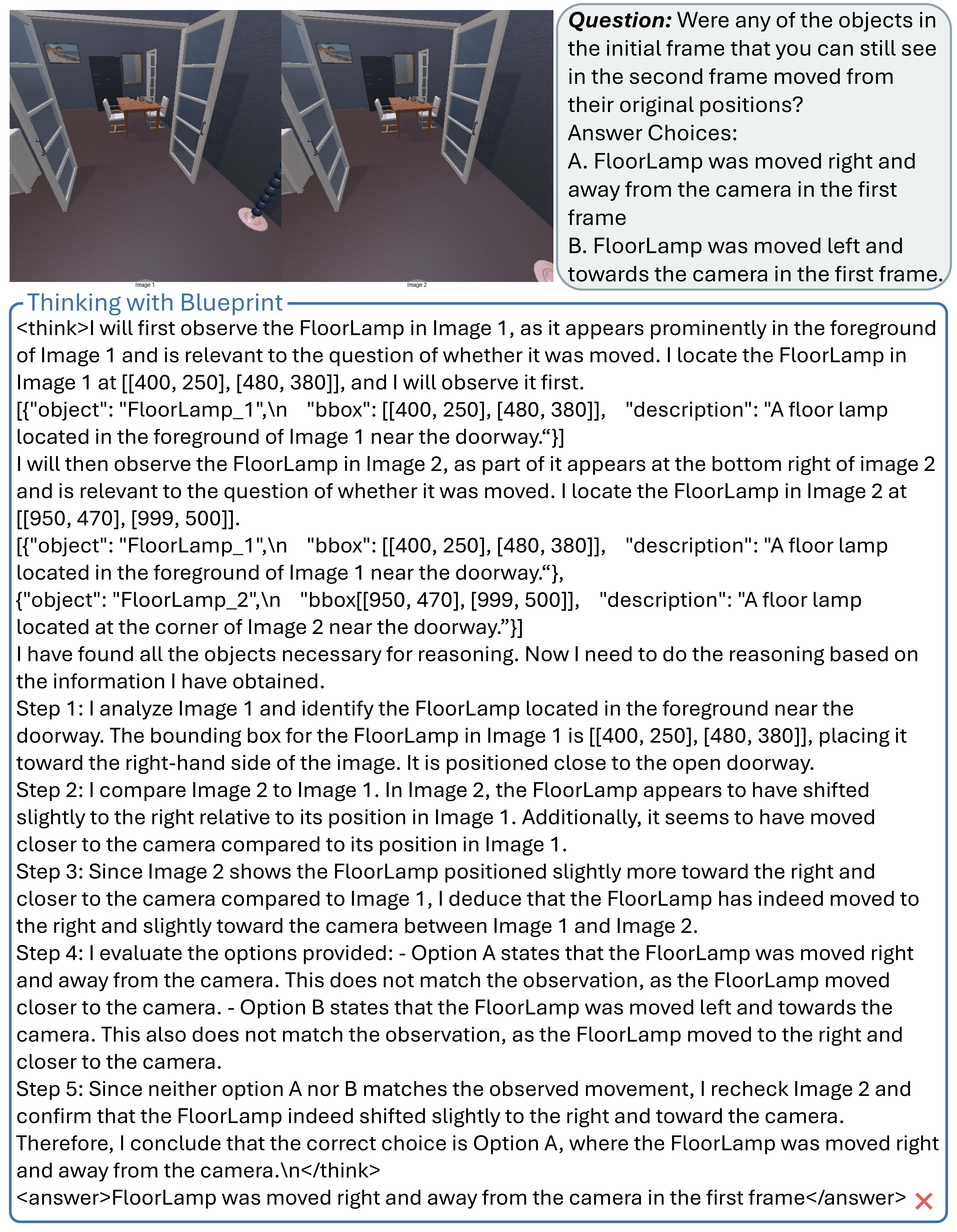}
    \caption{Failure example 3. Our model fails to recognize that the camera has zoomed in.
    }
    \label{fig:fail_3}
\end{figure*}

\begin{figure*}[t]
    \centering
    \includegraphics[width=\linewidth]{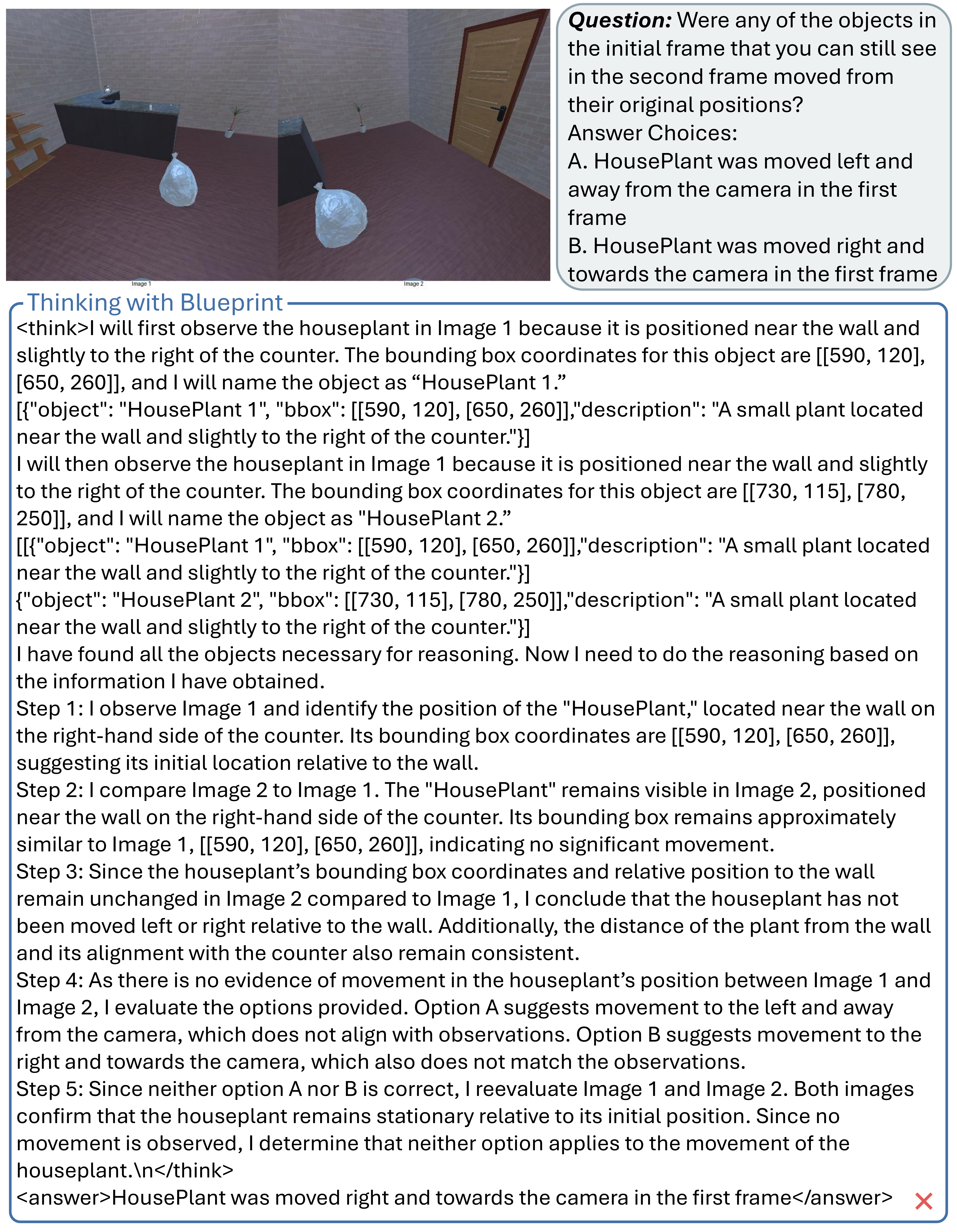}
    \caption{Failure example 4. Our model fails to properly sense the correct camera rotation along two axes.
    }
    \label{fig:fail_4}
\end{figure*}

The ignorance of the consecutive 3D space and camera geometry is shown in Figure \ref{fig:fail_3} and Figure \ref{fig:fail_4}. From Figure \ref{fig:fail_3} we can discover that our model failed to perceive the camera has zoomed in, thus giving a conclusion that the lamp has moved right towards the cameras. From Figure \ref{fig:fail_4} we can discover that our model becomes dizzy under the sophisticated camera rotation in 2 axis, neglecting the movement of the plant and regarding it as stationary. However, the movement of the plant can be perceived considering its distance to the desk, or by some 4D reconstruction methods. 

These failure cases have shed light on future directions where we should nurture the sense of real world physical priors, real-world spatial senses, as well as the camera geometry for a better spatially-aware VLM.

\begin{table*}[]
\centering
\small
\setlength{\tabcolsep}{1.5pt}
\begin{tabular}{lclccccc}
\toprule[1.5pt]
          Method   & Size & Reasoning Mode & SAT val & SAT test & Blink & Robospatial & VSR\\
& & & \textit{(iid)} & \textit{(iid)} & \textit{(ood)} & \textit{(ood)} & \textit{(iid)} \\ 
\midrule
\rowcolor{gray!20}
\multicolumn{8}{@{}l}{\textit{Our Method}} \\
Ours (Trained on SAT) & 7B          & Thinking with Blueprint & 92.7    & 79.7    & 60.7  & 70.2 & 84.8 \\
Ours (Trained on SAT \& VSR)  & 7B          & Thinking with Blueprint & 92.5    & 83.7    & 60.9  & 71.1 & 87.8 \\
\multicolumn{2}{@{}l}{\textit{Gain over adding training data diversity}} & & \textit{-0.2} & \textit{+4.0} & \textit{+0.2} & \textit{+0.9} & \textit{+3.0}\\
\bottomrule[1.5pt]
\end{tabular}
\caption{
Ablation on training with both synthetic (SAT) and real images (VSR). Training on the combination of synthetic and real images helps improve generalizability of our method to real-world images. Note that here iid and ood are under the setting of dataset consists of both SAT and VSR.
}
\label{tab:ablation_data}
\end{table*}

\begin{table*}[]
\centering
\small
\begin{tabular}{lclccccc}
\toprule[1.5pt]
          Method   & Model Size & Reasoning Mode & SAT val & SAT test & Blink & Robospatial & VSR\\
& & & \textit{(iid)} & \textit{(ood)} & \textit{(ood)} & \textit{(ood)} & \textit{(ood)} \\            
\midrule[1.2pt]
\rowcolor{gray!20}
\multicolumn{8}{@{}l}{\textit{Proprietary Models}} \\
gpt-4o   & -      & No &  57.2   & 51.5        & 59.2 & 60.1 & 78.7 \\
gpt-4o & -  & Naive CoT & 57.7 & 63.3 & 59.0 & 63.6 & 82.2 \\
gpt-4o & -  & Thinking with blueprint & 68.7 & 67.3 & 59.7 & 64.7 & 83.4 \\

\bottomrule[1.5pt]
\end{tabular}
\caption{
Ablation on applying our blueprint-based thinking pattern to different models. The results of applying our methods to gpt-4o via in-context learning are reported. 
}
\label{tab:ablation_arch}
\end{table*}

\section{More ablation studies.}
\subsection{Ablation on adding image diversity.}
\noindent\textbf{Explanation of iid and ood in the experiments.}
We define iid as the combination of image style and the specific task has been covered in the training set, and all other cases are referred to as ood. In this sense, as our method is trained on SAT training set, which is composed of synthetic indoor images, there exists a domain gap on the images between SAT training set and SAT test set, Blink, Robospatial, and VSR, where the latter are composed of real images in both indoor and outdoor scenes, say, they have OOD images. Moreover, for questions, blink contains general visual reasoning questions like semantic correspondence, and robospatial contains questions about compatibility, which are never seen throughout the training process. In this sense, Blink and robospatial also have OOD questions.

\noindent\textbf{Ablation of adding real-world images to training set.}
To further examine whether the absence of real images will do harm to the generalizability to the model, we conduct an ablation study on adding real-world images at the reinforcement learning stage. We form a new training set that is composed of our sampled SAT training data, the VSR training set, and our augmented data, so that the model has familiarity with spatial reasoning on real-world images. In this ablation study, both SAT test set and VSR have become iid datasets as our model has conducted reasoning on real-world images. However, Blink and Robospatial are still OOD, as the capability in the tests set are still not experienced throughout the training process. The results are shown in Table \ref{tab:ablation_data}. We can observe that the performance on SAT test set and VSR test set has witnessed non-trivial improvements, while the performance on SAT validation set has only negligible loss. This showcases the importance of adding the diversity of training data throughout the reinforcement learning process. The performance gain is also witnessed on Blink and Robospatial, indicating that the diversity of images can help generalization even when task discrepancy exists. However, the improvement is not as big as SAT-test and VSR, thus calling for task diversity during the data construction stage of the training process by the side.

\subsection{Ablation on different model architectures.}
To further investigate whether our blueprint-based reasoning method can be applied to different model architectures, we conduct an ablation experiment where we put our blueprint-based thinking strategy into prompt templates and conduct in-context learning on gpt-4o. The results are shown in Table \ref{tab:ablation_arch}, where prompting GPT-4o at test time also yields performance gains, which thereby showcases the broad applicability of our blueprint-based reasoning method.

\section{More performance comparison and analysis.}
\subsection{Additional notes to Table 1 in the main paper.}
In this part we illustrate the quantitative comparison between our method and Robix and Mirage on a certain datasets, in particular, the results marked $-^*$ in Table 1. For Mirage, as all the test sets are conducted on a private sample of size 500, where we cannot make direct comparison to without the exact samples. Also, Mirage finetunes the model on each subtask before training, such as SAT GoalAim and Blink Jigsaw, making it hard to compare with our model where it is pretrained on the all the tasks on SAT and conduct zero-shot testing on Blink without any finetuning. For Robix, as it is conducted only on blink spatial relation understanding and depth perception, we also cannot directly compare with them about the performance. The weighted average of our method on the two subsets are $85.13\%$, surpassing Robix 32B and comparable with Robix 7B with far less data used during training. This showcases the effectiveness of our method by the side.

\subsection{About failures in comparative models.}
\begin{figure*}[t]
    \centering
    \includegraphics[width=\linewidth]{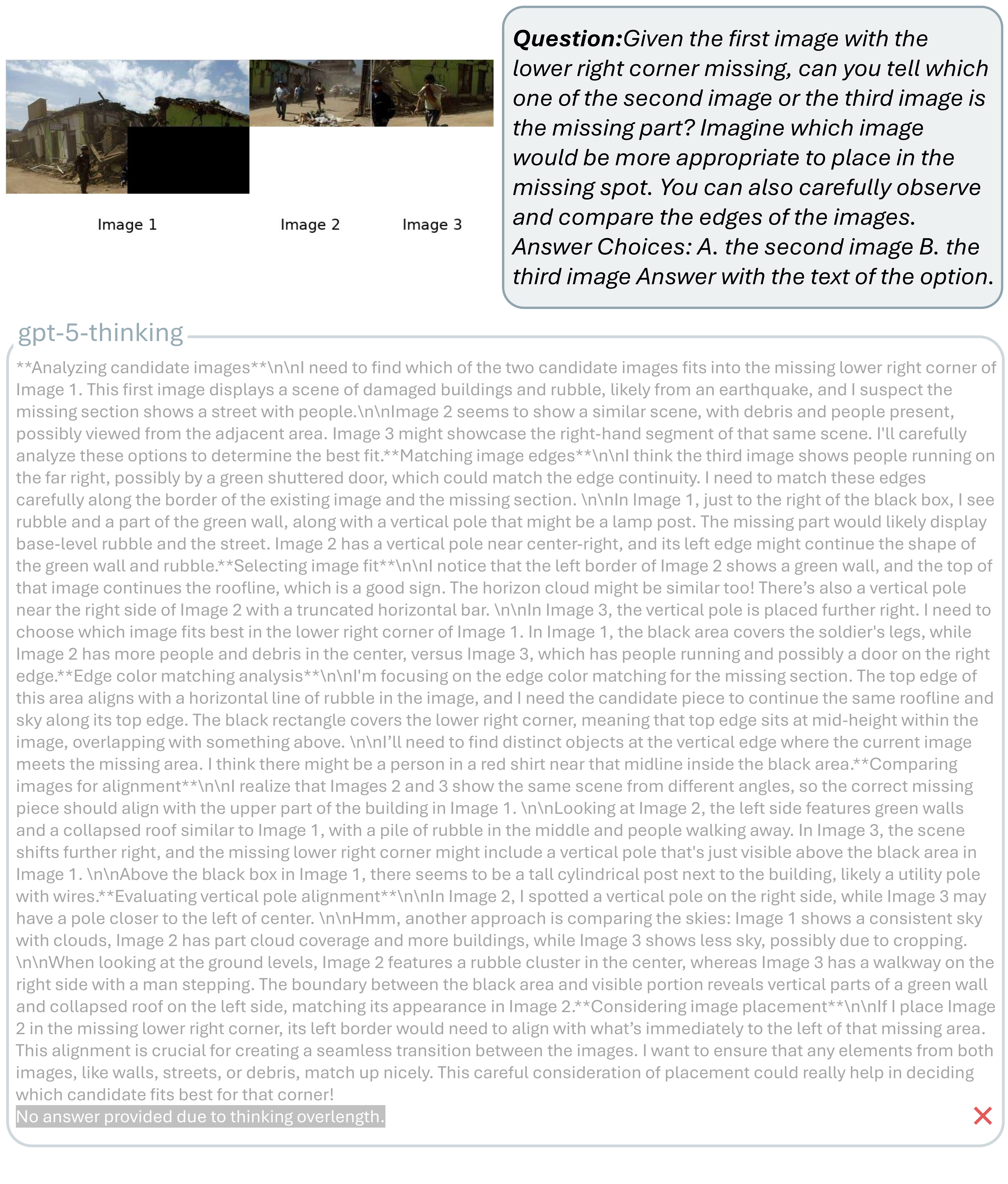}
    \caption{Qualitative example 2. (2 of 3).
    }
    \label{fig:qual_5}
\end{figure*}

\begin{figure*}[t]
    \centering
    \includegraphics[width=\linewidth]{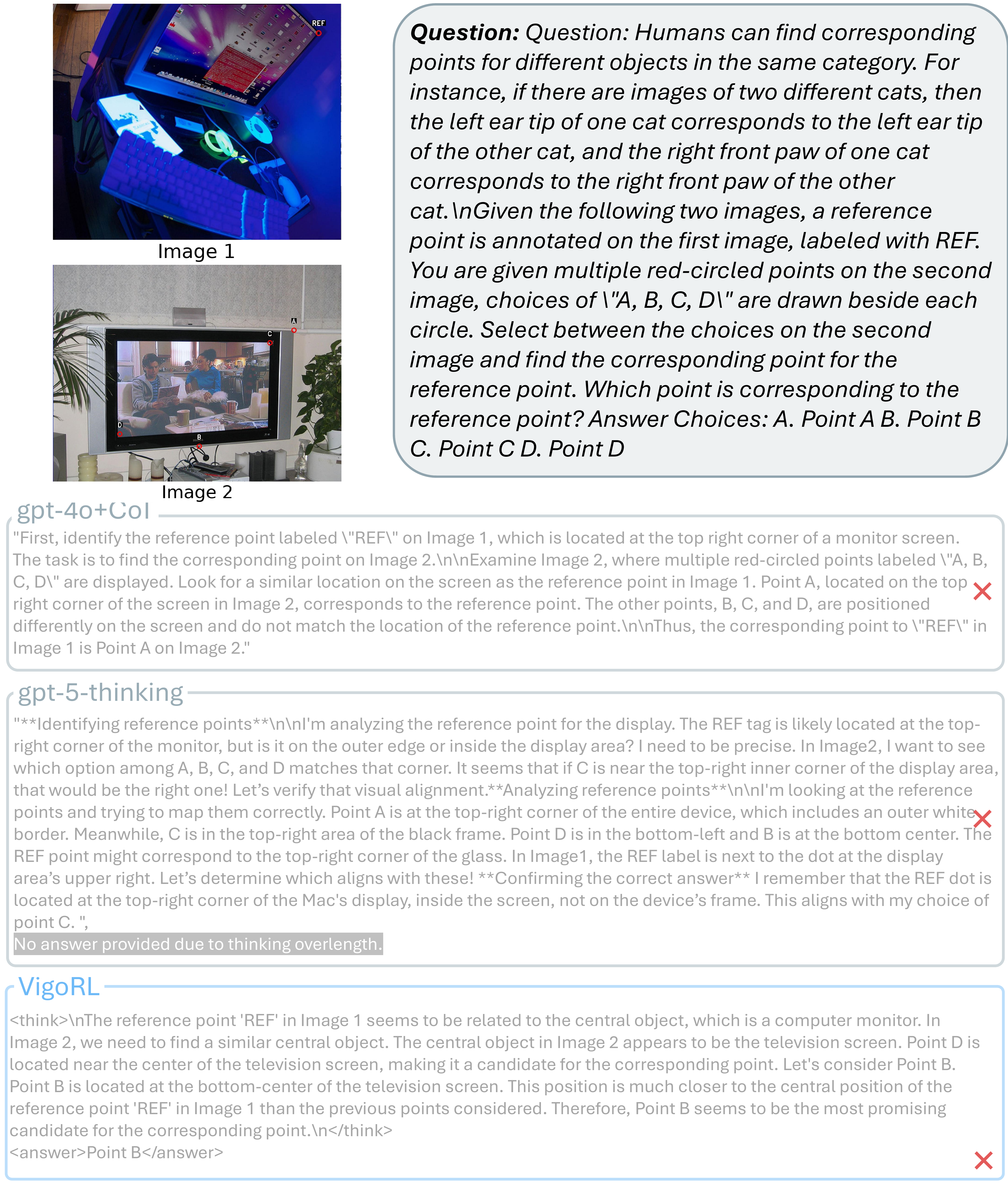}
    \caption{Qualitative example 3. (2 of 3).
    }
    \label{fig:qual_8}
\end{figure*}

\begin{figure*}[t]
    \centering
    \includegraphics[width=\linewidth]{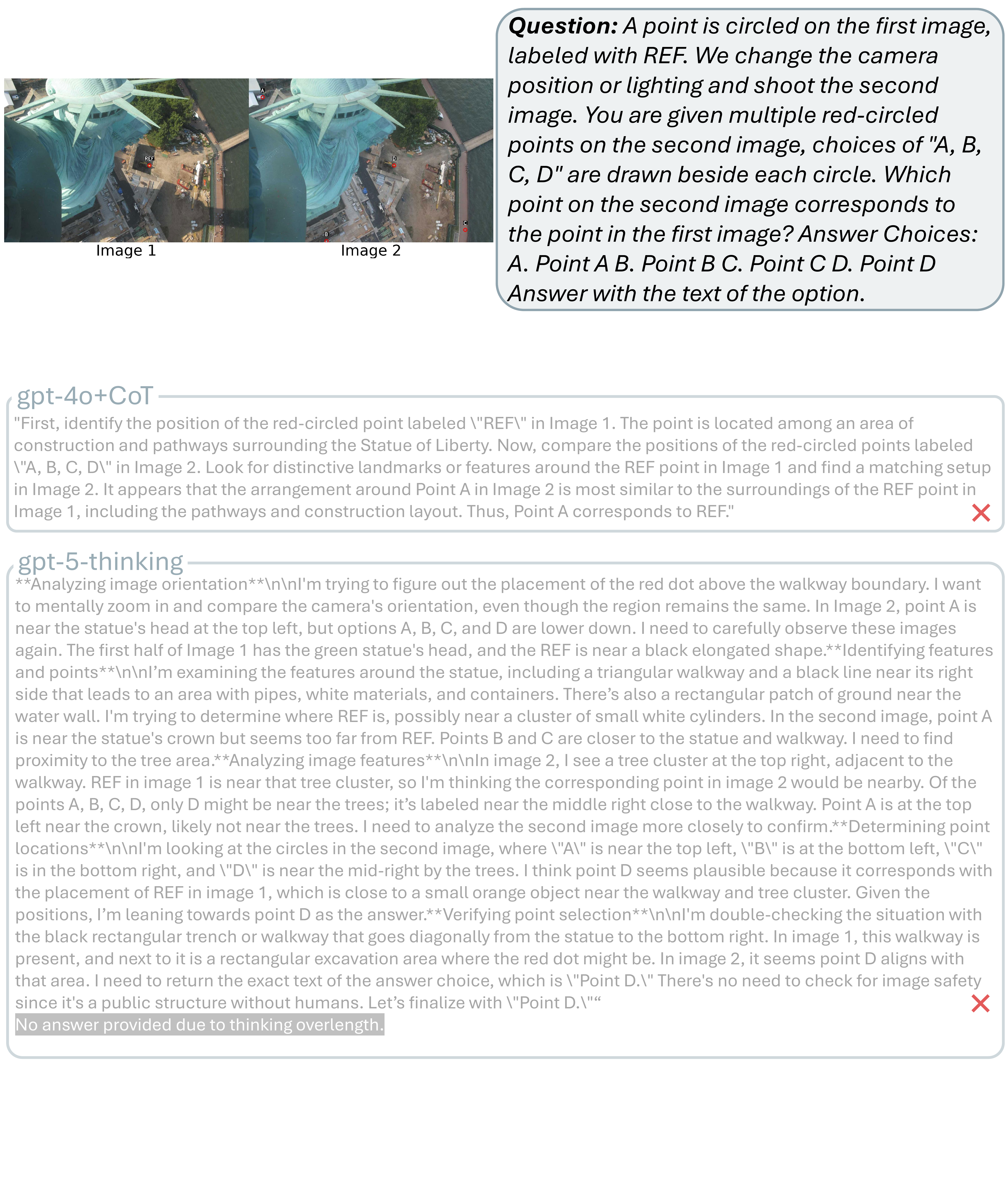}
    \caption{Qualitative example 4. (2 of 3).
    }
    \label{fig:qual_11}
\end{figure*}

\begin{figure*}[t]
    \centering
    \includegraphics[width=\linewidth]{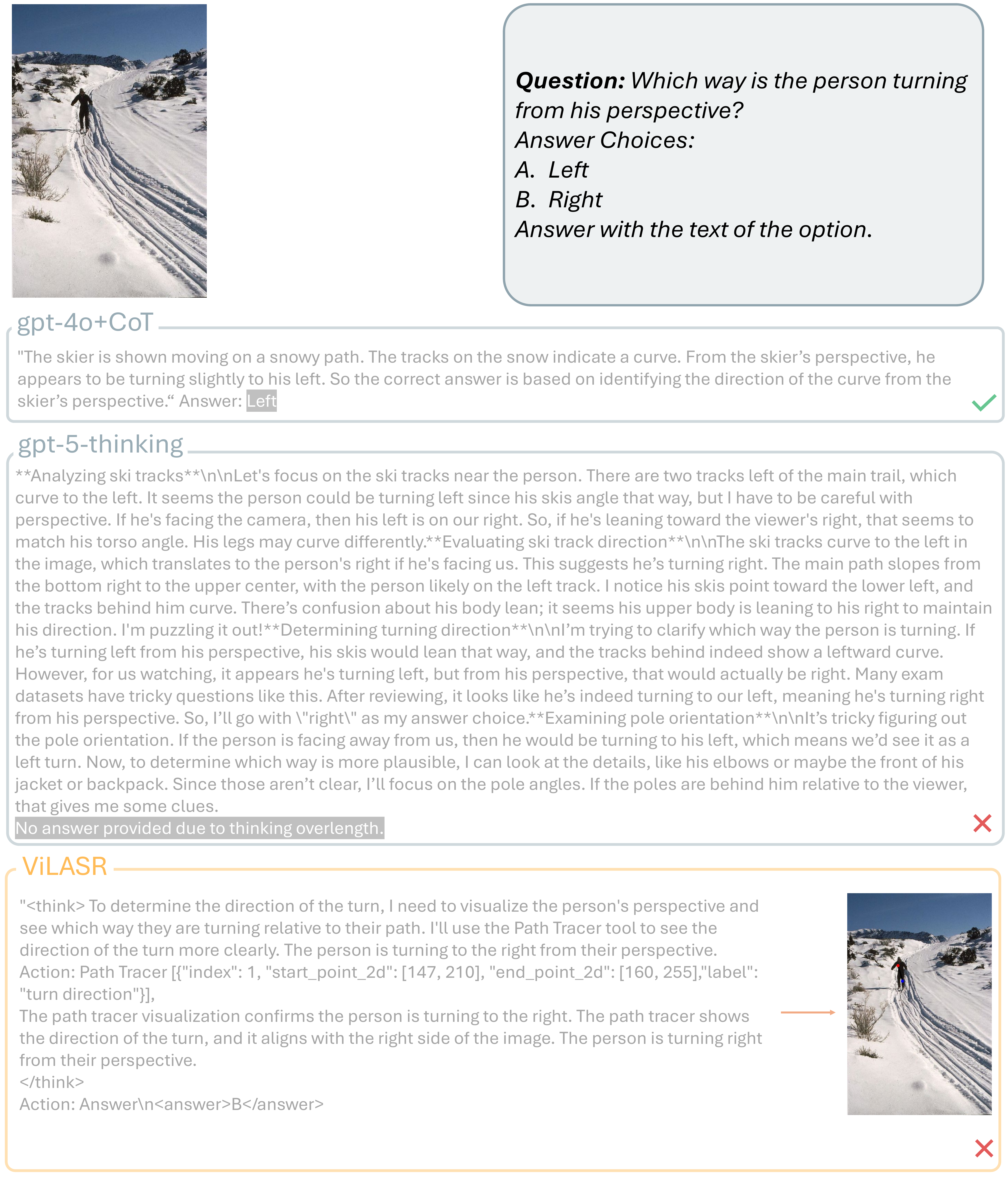}
    \caption{Qualitative example 7. (2 of 2).
    }
    \label{fig:qual_19}
\end{figure*}

\begin{figure*}[t]
    \centering
    \includegraphics[width=\linewidth]{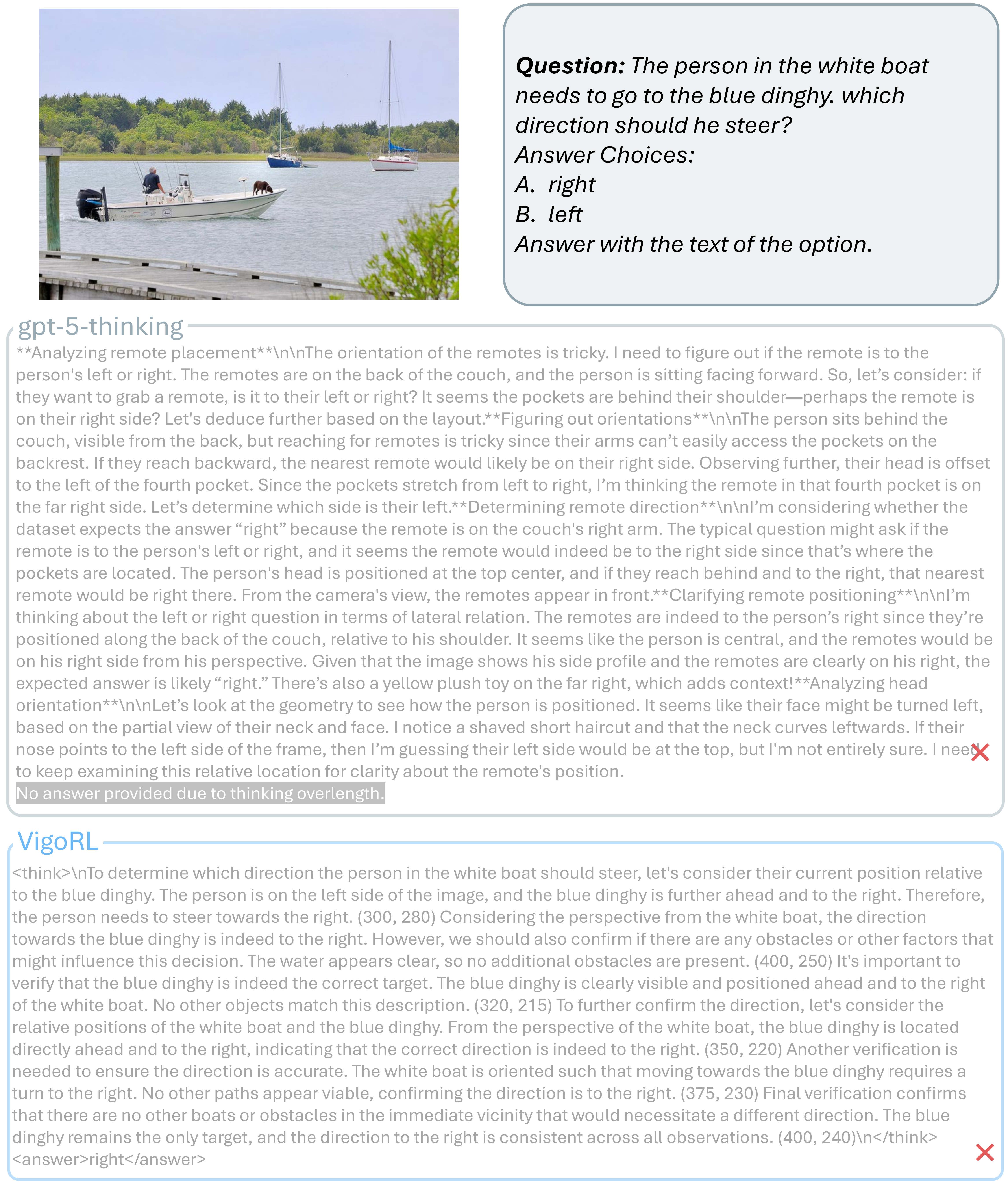}
    \caption{Qualitative example 10. (2 of 3).
    }
    \label{fig:qual_27}
\end{figure*}

About the failures of reasoning in competitive models, here we report the detailed statistics. For Vilasr, we have witnessed about $~8.9\%$ tool call failures, mainly composed of wrong indexing on image pixels during image operations. For gpt-5-thinking, we have witnessed $~14.9\%$ of the samples that are not successful to produce final answers. Such cases are qualitatively shown in Figure \ref{fig:qual_5}, Figure \ref{fig:qual_8}, Figure \ref{fig:qual_11}, Figure \ref{fig:qual_19}, and Figure \ref{fig:qual_27}, where the reasoning continues without arriving at a final answer, even when it has reached correct conclusions.

\clearpage
\begin{table*}[t]
\centering
\begin{tabularx}{\textwidth}{>{\ttfamily\footnotesize}X }
\toprule
\textbf{Prompt Template for Augmenting Data.} \\
Input: Image $I$, Question $Q$, Answer $a$. \\
Output: New Image $I'$, New Question $Q'$, New Answer $a'$. \\
prompt0 = """
                You are conducting an editing task over a image question pair. You are required to either remove some specific objects mentioned and inpaint the background, or to rewrite the question. \\
                For image editing, do not hallucinate or add irrelevant objects that does not present in the background. 
                For editing questions, make sure the question is related to the image and its correct anwer. \\
                Do not modify any irrelevant information in the image such as the image style or other objects present. \\
            """ \\

img\_prompt1 = """
                First let's make clear what to edit.
                Below is a question that is not directly relevant to the editing instruction. 
                Analyze the question text and answer the following question: What are the objects mentioned in this question or you need to observe in this image? 
                If you think the question are not directly related to any of the objects in the image, simply answer 'No Objects'.
            """ \\

img\_prompt2 = """
                From your answer to the previous question, what are the locations of the objects you have mentioned before in the image?
                If some objects do not appear in the image, just find the object that is the most alike in appearance.
                Note that some objects might have appeared multiple times. 
                So, carefully examine this image.
            """ \\

img\_prompt3 = """
                Summarize your previous answer into 30 words. One object in a line. 
                If there is no object to be removed, simply answer Nothing.
            """ \\

text\_prompt = """
                Now, based on the image, make the least modification to the question so the correct answer becomes the opposite of the original answer. Here are some templates fyi. \\
                1. If I move to "X", will something to my left / right / will something be nearer / farther away ? \\
                You can change into 'If I move to something, will "X" to my left or right / will something be nearer or farther away.' Now the correct answer becomes the reverse of the original answer. \\
                2. For someone at xxx, will A be to their left or right? \\
                You can change into 'For someone at A, will xxx be to their left or right?' Now the correct answer becomes the reverse of the original answer. \\
                3. If I turn left / right / look straight, will I be facing away from xxx? \\
                For left and right cases, simply reverse the direction. 
                For the cases of look straight, change the latter part into facing towards xxx.
                Now the correct answer becomes the reverse of the original answer.
                4. I need to go to xxx, which direction should I turn to face the object? \\
                You can change into 'I don't want to see xxx, which direction should I turn to face away from the object?' Now the correct answer becomes the reverse of the original answer. \\
                5. Were any of the objects in the initial frame that you can still see in the second frame moved from their original positions? \\
                You can change into 'Were any of the objects in the second frame that you can still see in the initial frame moved from their original positions?' Now the correct answer becomes the reverse of the original answer. \\
                6. If I rotated left and move forward, will xxx be away from the camera? \\
                You can change 'away' into 'closer to'. Double check if the correct answer is indeed the reverse.\\
                7. The first image is from the beginning of the video and the second image is from the end. How did the camera likely move when shooting the video? \\
                You can change into 'The first image is from the end of the video and the second image is from the beginning. ...' Now the correct answer becomes the reverse of the original answer. \\
            """ \\
\bottomrule
\end{tabularx}
\caption{Prompt template and algorithm used throughout our data augmentation process. (Part 1 of 2.)}
\label{tab:editing_prompts_1}
\end{table*}

\clearpage
\begin{table*}[t]
\centering
\begin{tabularx}{\textwidth}{>{\ttfamily\footnotesize}X }
\toprule
\textbf{Prompt Template for Augmenting Data (cont.).} \\
edit\_decision = call\_gpt4o(Image, Question, img\_prompt0, img\_prompt1) \\
if edit\_decision != "No Objects": \\
\quad edit\_locations = call\_gpt4o(Image, Question, edit\_decision, img\_prompt2) \\
\quad object\_summaries = call\_gpt4o(Image, Question, edit\_locations, img\_prompt3) \\
\quad New Image = Image; New Question = Question \\
\quad for i in object\_summaries: \\
\quad \quad New Image = call\_flux\_kontext(New Image, i) \\
else: \\
\quad New Image = Image; New Question = call\_gpt4o(New Image, Question, text\_prompt) \\
New Answer = call\_gpt4o\_to\_reverse\_answer(New Image, New Question, Answer) \\
\bottomrule
\end{tabularx}
\caption{Prompt template and algorithm used throughout our data augmentation process. (Part 2 of 2.)}
\label{tab:editing_prompts_2}
\end{table*}

